\documentclass[runningheads]{llncs}
\usepackage[T1]{fontenc}
\usepackage{graphicx}
\usepackage{booktabs}
\usepackage[misc]{ifsym}
\newcommand{\corr}{(\Letter)}
\usepackage{mwe}

\usepackage{amsmath}
\usepackage{amssymb}
\usepackage{multirow}
\usepackage{pifont}
\usepackage{tabularx}
\usepackage[table,xcdraw]{xcolor}

\usepackage{enumitem}
\usepackage{bbding}

\begin{document}

\title{Revisiting Applicable and Comprehensive Knowledge Tracing in Large-Scale Data}

\titlerunning{DKT2}

\author{Yiyun Zhou   \and
Wenkang Han\and
Jingyuan Chen \corr}

\authorrunning{Y. Zhou et al.}

\institute{Zhejiang University \email{\{yiyunzhou, wenkangh, jingyuanchen\}@zju.edu.cn}}

\maketitle              
\begin{abstract}
\label{sec:abstract}
\begin{sloppypar}
Knowledge Tracing (KT) is a fundamental component of Intelligent Tutoring Systems (ITS), enabling the modeling of students' knowledge states to predict future performance. The introduction of Deep Knowledge Tracing (DKT), the first deep learning-based KT (DLKT) model, has brought significant advantages in terms of applicability and comprehensiveness. However, recent DLKT models, such as Attentive Knowledge Tracing (AKT), have often prioritized predictive performance at the expense of these benefits. While deep sequential models like DKT have shown potential, they face challenges related to parallel computing, storage decision modification, and limited storage capacity. 
To address these limitations, we propose DKT2, a novel KT model that leverages the recently developed xLSTM architecture. DKT2 enhances applicable input representation using the Rasch model and incorporates Item Response Theory (IRT) for output interpretability, allowing for the decomposition of learned knowledge into familiar and unfamiliar knowledge. By integrating this knowledge with predicted questions, DKT2 generates comprehensive knowledge states. Extensive experiments conducted across three large-scale datasets demonstrate that DKT2 consistently outperforms 18 baseline models in various prediction tasks, underscoring its potential for real-world educational applications. This work bridges the gap between theoretical advancements and practical implementation in KT. Our code and datasets are fully available at \url{https://github.com/zyy-2001/DKT2}.
\end{sloppypar}

\end{abstract}

\section{Introduction}
\label{sec: introduction}

\begin{sloppypar}

The rapid expansion of educational data within Intelligent Tutoring Systems (ITS)~\cite{luckin2016intelligence} (\textit{e.g.}, AutoTutor~\cite{nye2014autotutor}) has exposed significant limitations in traditional machine learning approaches~\cite{bengio2013representation}. In contrast, the advent of deep learning has introduced novel opportunities for addressing these challenges~\cite{lecun2015deep,wu2025embracingimperfectionsimulatingstudents}. A critical component of ITS is Knowledge Tracing (KT), which models students' knowledge states and predicts future performance by analyzing their interaction data. Deep learning, with its advanced feature learning paradigm, offers enhanced modeling power and predictive accuracy in this context.

Deep Knowledge Tracing (DKT)~\cite{NIPS2015_bac9162b} represents the first significant application of deep learning to KT, employing Long Short-Term Memory (LSTM) networks~\cite{hochreiter1997long} to capture the complexity of students' learning processes. As a pioneering deep learning-based KT (DLKT) model, DKT has demonstrated superior predictive performance compared to traditional machine learning-based KT models (\textit{e.g.}, Bayesian Knowledge Tracing (BKT)~\cite{corbett1994knowledge}), offering notable advantages in applicability and comprehensiveness.

DKT encodes students' \textbf{historical} interactions to generate a comprehensive representation of their knowledge states (\textit{i.e.}, proficiency scores\footnote{Proficiency scores range from 0 to 1, with higher values indicating greater knowledge and skill level.} for \textbf{each concept} at each time step) and predicts future performance. \textbf{However, recent DLKT models, such as the Attentive Knowledge Tracing (AKT)~\cite{ghosh2020context}, while excelling in predictive accuracy~\cite{liu2023simplekt,im2023forgetting,huang2023towards,yin2023tracing,zhou2025cuffkttacklinglearnersrealtime}, present limitations in applicability and comprehensiveness (\textbf{the related details are in Sec.~\ref{sec: akt}}).} Specifically, AKT requires both historical and future interactions as input, complicating its practical application since future responses are typically unavailable. Additionally, unlike DKT, AKT directly predicts scores on future questions without generating a comprehensive knowledge state, potentially weakening the correlations between different concepts and narrowing the definition of knowledge states in KT. Our review of 60 KT-model-related papers published in top AI/ML conferences and journals over the past decade (see Appendix~\ref{apx: apx-summary}) reveals a trend where evaluation performance has been prioritized at the expense of practical applicability, risking a disconnect between theoretical advancements and real-world implementation.

Deep sequential models like DKT have intrinsic limitations that may prevent them from achieving optimal performance. LSTM networks, for instance, face challenges in dynamically updating stored information and exhibit limited storage capacity due to their scalar cell state design. Moreover, their inherent sequential processing nature hinders parallelization, limiting their scalability to large datasets. The recently proposed xLSTM~\cite{beck2024xlstm}, however, addresses these challenges by introducing two new variants: sLSTM, which improves LSTM's storage decision by incorporating an exponential activation function, and mLSTM, which replaces scalar cell states with matrix memory for increased storage capacity and improved retrieval efficiency, while achieving full parallelization by abandoning memory mixing. Building on the strengths of xLSTM, we introduce DKT2, an enhanced DLKT model designed for greater applicability and comprehensiveness. DKT2 integrates the Rasch model~\cite{rasch1993probabilistic} from educational psychology to process historical interactions, using xLSTM for knowledge learning. DKT2 then incorporates Item Response Theory (IRT)~\cite{lord1952theory,yen2006item} to interpret the learned knowledge, differentiating between familiar and unfamiliar knowledge, and ultimately integrates this knowledge with predicted questions to generate comprehensive knowledge states.

Our primary contributions are as follows:
\begin{itemize}[leftmargin=*]
    \item{We provide a systematic analysis of input and output settings in KT, proposing DLKT models optimized for real-world applicability and comprehensiveness.}
    \item{We introduce DKT2, a model built on xLSTM, adhering to rigorous applicable input and comprehensive output settings, and incorporating both the Rasch model for input and an interpretable IRT-based output module.}
    \item{We conduct extensive experiments, including one-step prediction, multi-step prediction, and predictions with varying history lengths, across three large-scale datasets. Our findings demonstrate that DKT2 consistently outperforms 18 baseline models, with additional analysis on the impact of input settings and multi-concept output predictions on KT performance.}
\end{itemize}

\end{sloppypar}

\begin{sloppypar}
\section{Related Work}
\label{sec: related_work}

Since DKT~\cite{NIPS2015_bac9162b} first applied deep learning methods to the KT task a decade ago, deep learning techniques have flourished in KT. Current DLKT models can be categorized into the following 8 types:
\begin{itemize}[leftmargin=*]
\item{\textbf{Deep sequential models} use recurrent structures to encode students' chronologically ordered interactions, \textit{e.g.}, DKT uses LSTM to model complex student cognitive processes. Two variants of DKT have emerged in subsequent research. DKT+~\cite{yeung2018addressing} introduces two regularization terms to improve the consistency of KT predictions, while DKT-F~\cite{nagatani2019augmenting} enhances KT by considering forgetting behavior.}
\item{\textbf{Attention-based models} capture long-term dependencies between interactions through attention mechanisms, \textit{e.g.}, SAKT~\cite{pandey2019self} is the first to use attention mechanisms to capture correlations between concepts and interactions. AKT~\cite{ghosh2020context} employs a novel monotonic attention to represent the time distance between questions and students' historical interactions. Due to AKT's outstanding predictive performance, numerous powerful KT models are subsequently derived, such as simpleKT~\cite{liu2023simplekt}, FoLiBiKT~\cite{im2023forgetting}, sparseKT~\cite{huang2023towards}, DTransformer~\cite{yin2023tracing}, and stableKT~\cite{li2024enhancing}.}
\item{\textbf{Mamba-based models} are strong competitors to Transformer models. The recently proposed Mamba4KT~\cite{cao2024mamba4kt} is the first KT model to explore evaluation efficiency and resource utilization.}
\item{\textbf{Graph-based models} use graph structures to characterize the relationships between questions, concepts, or interactions, \textit{e.g.}, GKT~\cite{nakagawa2019graph} uses a graph to model the intrinsic relationships between concepts.}
\item{\textbf{Memory-augmented models} capture latent relationships between concepts through memory networks, \textit{e.g.}, DKVMN~\cite{zhang2017dynamic} uses a static key matrix to store relationships between different concepts and updates students' knowledge states through a dynamic value matrix. SKVMN~\cite{abdelrahman2019knowledge}, a variant of DKVMN, also integrates the advantages of LSTM in recurrent modeling.}
\item{\textbf{Adversarial-based models} use adversarial techniques to enhance the model's generalization ability, \textit{e.g.}, ATKT~\cite{guo2021enhancing} mitigates overfitting and improves generalization by adding perturbations to student interactions during training.}
\item{\textbf{Contrastive learning-based models} use contrastive learning to learn rich representations of student interactions, \textit{e.g.}, CL4KT leverages contrastive learning to strengthen representation learning by distinguishing between similar and dissimilar learning histories.}
\item{\textbf{Other representative models} include interpretable models and models with auxiliary tasks, \textit{e.g.}, Deep-IRT~\cite{yeung2019deep} introduces item response theory~\cite{lord1952theory} based on DKVMN to make deep learning-based KT explainable. AT-DKT~\cite{liu2023enhancing} enhances KT by introducing two auxiliary learning tasks: question tagging prediction and individualized prior knowledge prediction.}
\end{itemize}

Our proposed DKT2, by breaking the parallelization limitations of deep sequential models, can be classified as a new type of deep sequential models (\textbf{Deep sequential models$^\ast$}).

\end{sloppypar}

\begin{sloppypar}
\section{Methodology}
\label{sec:method}
\setlength{\abovedisplayskip}{3pt}
\setlength{\belowdisplayskip}{3pt}
\begin{figure*}[t]
    \includegraphics[width=\linewidth]{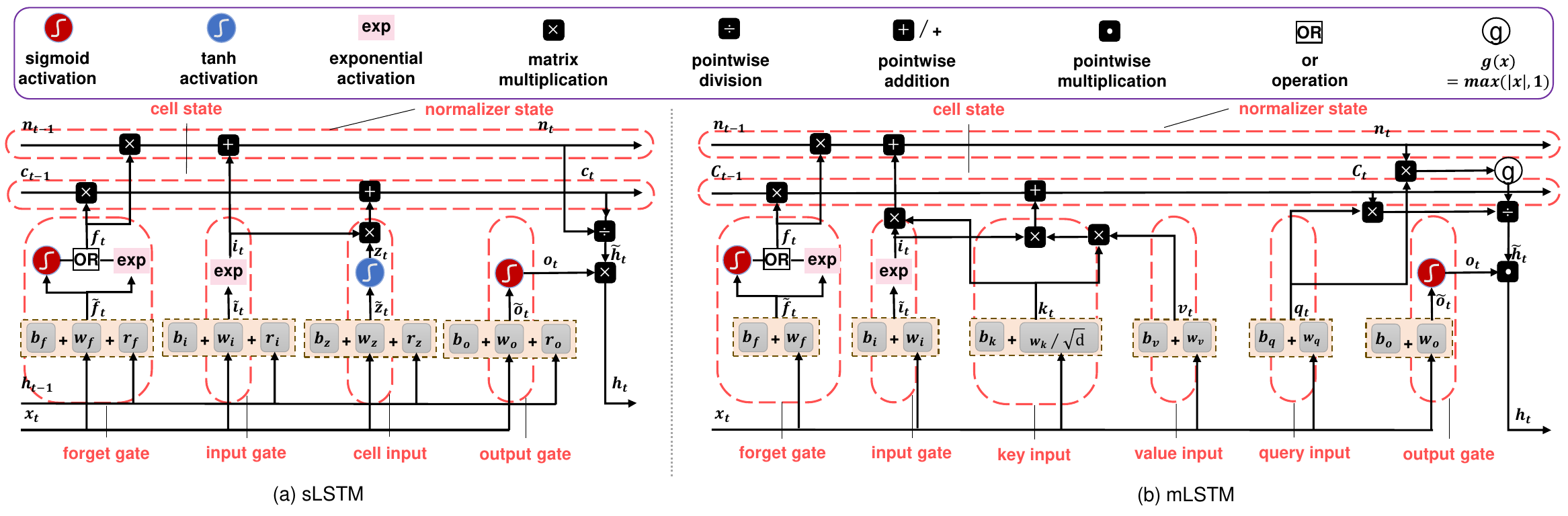}
    \vspace{-0.6cm}
    \caption{Architecture of xLSTM.}
    \label{fig:xlstm}
    \vspace{-0.5cm}
\end{figure*}

\subsection{Problem Statement}
\label{sec: statement}

In the KT task, formally, let $\mathcal{S}$, $\mathcal{Q}$, and $\mathcal{C}$ represent the sets of students, questions, and concepts respectively. For each student $s\in \mathcal{S}$, there exists a sequence of $k$ time steps $X_k = \{(q_1, c_1, r_1,t_1 ), (q_2, c_2, r_2,t_2 ),\ldots ,(q_k, c_k, r_k,t_k)\}$, where $q_i\in \mathcal{Q}, c_i\subset \mathcal{C}, r_i\in \{0, 1\}$, and $t_i$ represent the question attempted by the student, the concepts related to question $q_i$, whether the student responded correctly (0 for incorrect, 1 for correct), and the timestamp of the response, respectively. At time step $k+1$, DKT2 predicts $\hat{r}_{k+1}$ based on the student's interaction sequence $X_k$:
\begin{flalign}
\hat{r}_{k+1} = \text{DKT2}(X_k, q_{k+1}, c_{k+1}, t_{k+1}\mid\theta),
\end{flalign}
where $\theta$ represents the parameters learned during training.

\subsection{Preliminaries}
\label{sec: preliminary}

\subsubsection{LSTM and the Extended LSTM}
\label{sec: lstm}

LSTM\footnote{Refer to Appendix~\ref{apx: apx-lstm} for details on LSTM.} is one of the earliest popular deep learning methods applied to NLP, but it has been overshadowed for a period by the success of Transformers~\cite{vaswani2017attention,zhou2025cola}. However, the architecture is recently regaining attention and undergoing significant improvements. The improved LSTM is called extended Long Short-Term Memory (xLSTM)~\cite{beck2024xlstm}, which mainly addresses three limitations in traditional LSTM: (1) inability to revise storage decisions, (2) limited storage capacities, and (3) lack of parallelizability. xLSTM introduces two new members to the LSTM family to overcome these limitations: sLSTM and mLSTM, as described in Fig.~\ref{fig:xlstm}. \textbf{Since our work does not focus on the architecture of xLSTM, we have placed the detailed introduction of xLSTM in Appendix~\ref{apx: apx-xlstm}.}

\subsubsection{Weakly Applicable Input and Comprehensive Output Settings in DLKT Models}
\label{sec: akt}

\begin{figure}[t]
    \includegraphics[width=\linewidth]{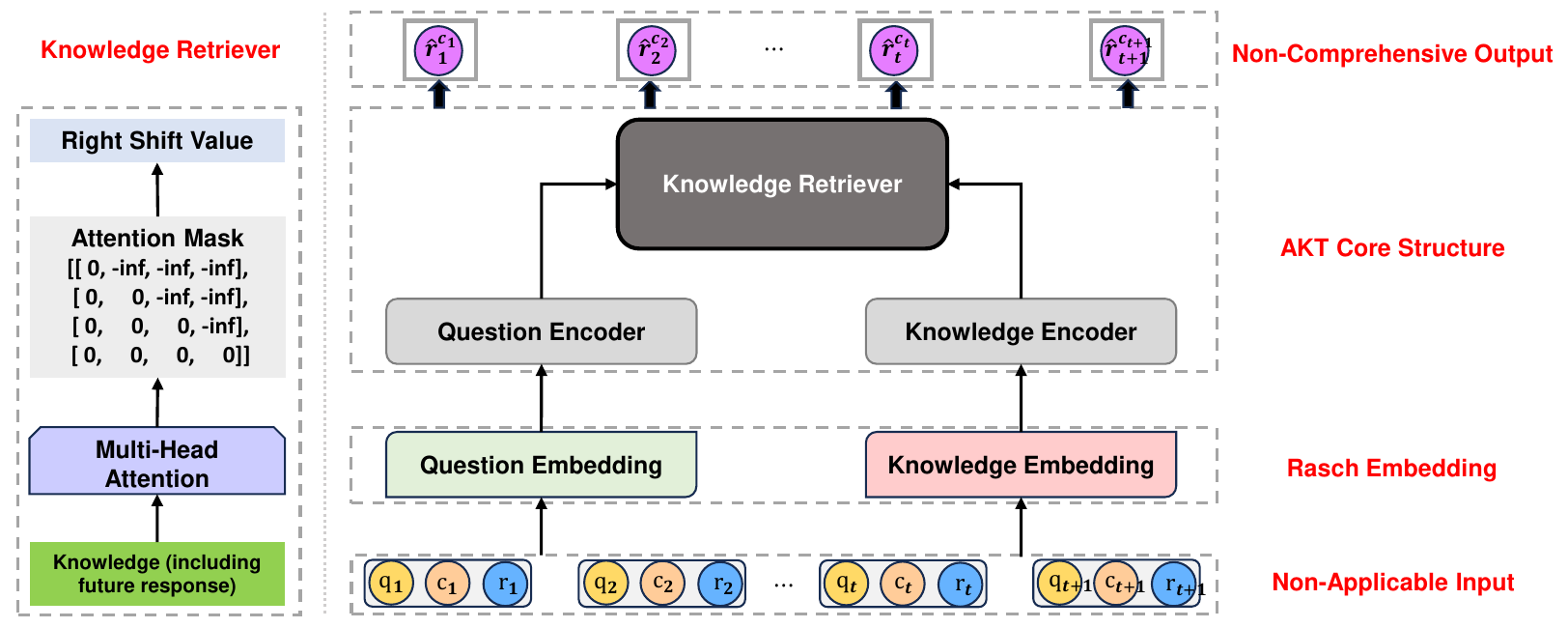}
    \vspace{-0.7cm}
    \caption{Structural sketch of AKT.}
    \label{fig: akt}
    \vspace{-0.5cm}
\end{figure}

We use AKT~\cite{ghosh2020context} as an example to describe the common weakly applicable input and comprehensive output settings in DLKT models. Fig.~\ref{fig: akt} shows a structural sketch of AKT. Clearly, AKT takes both historical interactions and future interactions as input during training and inference, ignoring future information through attention masking while representing knowledge learned up to the current time step through offset (right-shifting values in attention), and directly predicts questions at each time step. From this, we can see that although AKT's setup is reasonable and does not lead to future information leakage, this input setting, while convenient, also \textbf{causes complications in engineering implementation} (engineering often requires cumbersome representation of future information as a padding value, and this common processing method does not seem suitable for KT, as KT tasks typically involve predicting future questions $2\sim t+1$ based on historical interactions $1\sim t$). Moreover, AKT only outputs the response for the current time step's question, without considering the student's proficiency in different dimensions, which \textbf{contradicts the multidimensional nature of real-world student knowledge and narrows the definition of KT.}

\begin{figure*}[!htb]
\centering
    \includegraphics[width=0.7\linewidth]{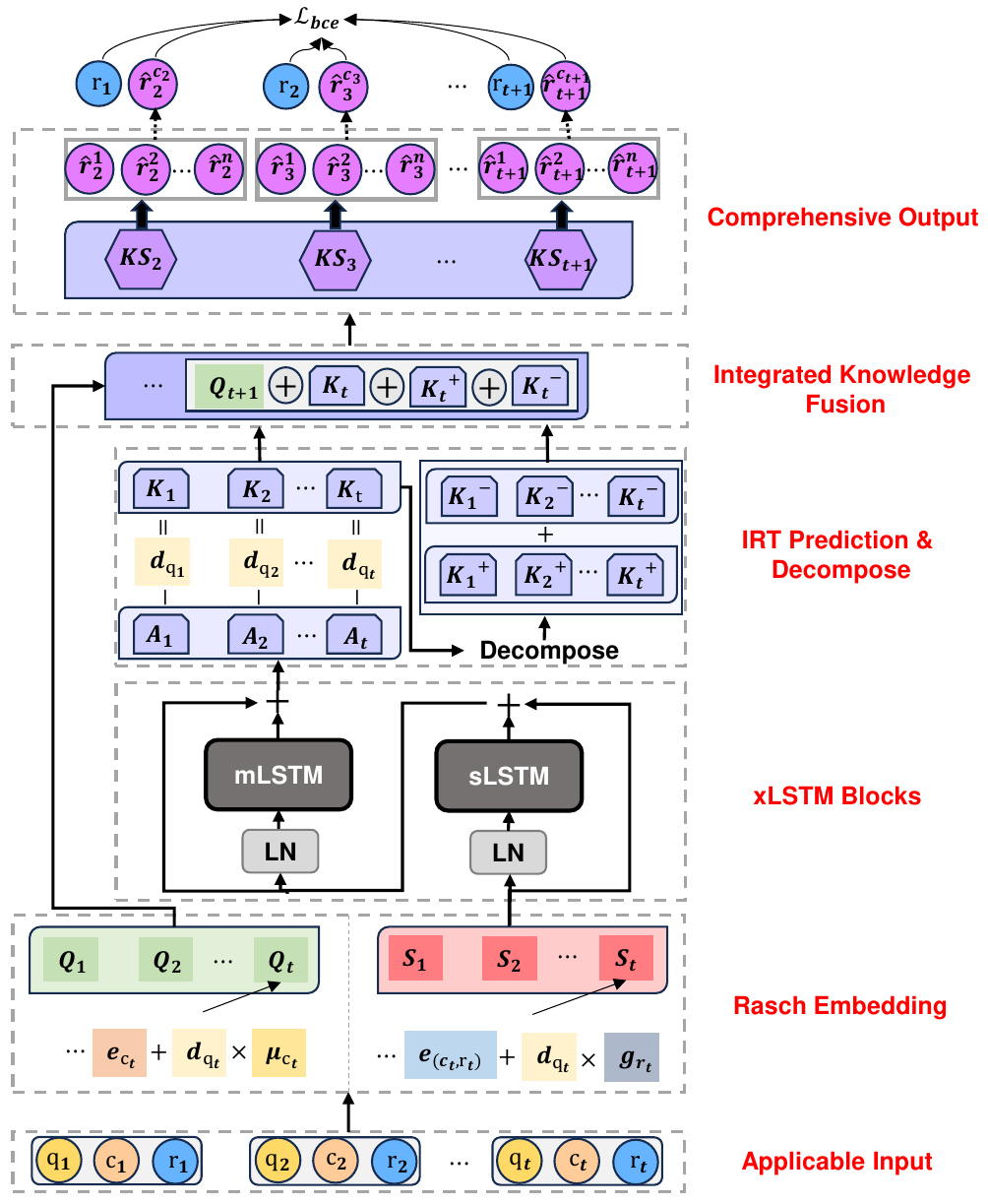}
    \vspace{-0.3cm}
    \caption{Architecture of DKT2.}
    \label{fig: dkt2}
    \vspace{-0.5cm}
\end{figure*}

\subsection{DKT2}
\label{sec: dkt2}
Fig.~\ref{fig: dkt2} illustrates the architecture of our proposed DKT2, as described below.

\subsubsection{Rasch Embedding}

We use the classic Rasch model~\cite{rasch1993probabilistic} from educational psychology to construct embeddings of questions and student skills. This model explicitly uses scalars to represent the degree of deviation between questions and the concepts they cover. Additionally, we choose to use question-specific difficulty vectors to capture differences among various questions within the same concept. DKT2 takes \textbf{applicable interactions (\textit{i.e.}, inputs not involving the future response $r_{t+1}$, distinguishing it from models like AKT)} as input, denoted as $\{q_i, c_i, r_i\}_{i=1}^t$, and at time step $t$, the embeddings of questions and student skills, $Q_t$ and $S_t$ respectively, are represented as:
\begin{equation}
    Q_t = e_{c_t} + d_{q_t} \cdot \mu_{c_t}, S_t = e_{(c_t, r_t)} + d_{q_t} \cdot g_{r_t}, e_{(c_t, r_t)} = e_{c_t} + e_{r_t},
\label{eq: embedding}
\end{equation}
where $e_{c_t}\in \mathbb{R}^d$ and $e_{r_t}\in \mathbb{R}^d$ are the embeddings of concept $c_t$ and response $r_t$, respectively. $d_{q_t}\in \mathbb{R}$ is a difficulty scalar and $\mu_{c_t}\in \mathbb{R}^d$ summarizes the variation of questions containing concept $c_t$. $e_{(c_t, r_t)}\in \mathbb{R}^d$ is the interaction representation of the concept and student response, $g_{r_t}\in \mathbb{R}^d$ is the variant embedding of the response. $d$ is the dimension of the embeddings.

\subsubsection{xLSTM Blocks}
DKT2 further learns the student's ability representation $A_{1:t}$ at time step $t$ through two xLSTM blocks (sLSTM and mLSTM) based on the original representation of student ability $S_{1:t}$:
\begin{equation}
\begin{aligned}
    A_{1:t} = \text{Res}\left(\text{LN}\left(\text{mLSTM}\left(\text{Res}\left(\text{LN}\left(\text{sLSTM}(S_{1:t})\right)\right)\right)\right)\right),
\end{aligned}
\end{equation}
where LN and Res refer to layer normalization~\cite{ba2016layer} and residual connection~\cite{He2015}, respectively.

\subsubsection{IRT Prediction \& Decompose}
The core idea of IRT (Item Response Theory) lies in the interactive relationship between student ability and question difficulty~\cite{yen2006item}. Specifically, \textbf{if a student's ability is far above the question's difficulty, the probability of the student responding to the question correctly is very high, and vice versa.} This is also why IRT is often used for interpretable predictions in KT~\cite{yeung2019deep,sun2024interpretable} (\textbf{our work focuses not on the interpretability of KT models but on evaluating their applicability and comprehensive setup}). Therefore, the knowledge acquired by a student, denoted as $K_{1:t}$, can be represented as:
\begin{equation}
\begin{split}
    K_{1:t} = A_{1:t} - d_{q_{1:t}},
\end{split}
\end{equation}
where $d_{q_{1:t}}$ is the sequence representation of $d_{q_t}$ from Eq.~\ref{eq: embedding} up to time step $t$.

Further, DKT2 roughly distinguishes between the familiar and unfamiliar knowledge $K_{1:t}^+$ and $K_{1:t}^-$ based on correct and incorrect responses:
\begin{equation}
\begin{split}
    K_{1:t}^{+} = \text{exp}(r_{1:t}, d)\circ K_{1:t}, K_{1:t}^{-} = \text{exp}(\textbf{one} - r_{1:t}, d)\circ K_{1:t},
\end{split}
\end{equation}
where $\text{exp}(\cdot, d)$ denotes expanding the last dimension of the tensor to $d$ dimensions. $\circ$ denotes element-wise multiplication. $\textbf{one}\in \mathbb{R}^t$ is a vector of all ones.

\subsubsection{Integrated Knowledge Fusion}

DKT2 estimates the student's knowledge $X_{2:t+1}$ based on the knowledge $K_{1:t}$ and the questions $Q_{2:t+1}$ that need to be predicted:
\begin{equation}
\begin{split}
    X_{2:t+1} = Q_{2:t+1} \oplus K_{1:t} \oplus K_{1:t}^{+} \oplus K_{1:t}^{-},
\end{split}
\end{equation}
where $\oplus$ denotes the concatenation operation. In addition to integrating the questions and the student's current knowledge, DKT2 also includes the  student's familiar and unfamiliar knowledge $K_{1:t}^{+}$ and $K_{1:t}^{-}$. This is because, intuitively, \textbf{if the knowledge required to respond to a question is familiar to the student, the predicted score tends to be higher, and conversely, lower if unfamiliar.}

Finally, DKT2 predicts the student's comprehensive knowledge states $\text{KS}_{2:t+1}$:
\begin{equation}
\begin{split}
    \text{KS}_{2:t+1} = \sigma(\text{ReLU}(X_{2:t+1}W_1+b_1)W_2+b_2),
\end{split}\label{eq: ks}
\end{equation}
where $W_1\in \mathbb{R}^{4d\times 2d},W_2\in \mathbb{R}^{2d\times n}, b_1\in \mathbb{R}^{2d},b_2\in \mathbb{R}^{n}$ are learnable parameters in the MLP. $\sigma(\cdot)$ is the Sigmoid function and $\text{ReLU}(\cdot)$ is the activation function. $n$ is the number of concepts (due to data sparsity, KT often predicts the concepts corresponding to questions).

Eq.~\ref{eq: ks} can be further represented as:
\begin{equation}
\begin{split}
\text{KS}_i = (\hat{r}_i^1, \hat{r}_i^2,\dots, \hat{r}_i^n), 2\leq i \leq t+1,
\end{split}
\end{equation}
where $\hat{r}_i^j$ represents the prediction score of DKT2 for concept $j$ at time step $i$. \textbf{The comprehensive output of DKT2 enables the prediction of multiple concepts at the same time step, whereas models like AKT can only predict $\hat{r}_i^{c_i}$ at time step $i$}. We will analyze the multi-concept prediction scenario in detail in Sec.~\ref{sec: output}, where some unexpected results have been discovered.

\subsubsection{Model Training}
The loss of DKT2 is defined as the binary cross-entropy loss between the prediction ${\hat{r}}_{t}$ and the actual response $r_t$, calculated as follows:
\begin{flalign}
    \mathcal{L}_{\text{DKT2}} = -\sum_{i=2}^{t+1}{r_{i}\text{log}({\hat{r}}_{i})+(1-r_{i})\text{log}(1-{\hat{r}}_{i})}.
\end{flalign}

\subsubsection{Conversion of Input and Output Settings}
We attempt to convert the weakly applicable input and comprehensive output settings in DLKT models into strongly applicable input and comprehensive output settings. Similarly, using AKT as an example, like DKT2, as shown in Fig.~\ref{fig: dkt2}, the transformed AKT only takes the historical interactions $\{(q_i, c_i, r_i)\}_{i=1}^t$ as input, with everything else remaining unchanged (note that the right-shift operation still needs to be retained because the attention does not mask the knowledge of the current time step). Before outputting the predicted score $\hat{r}_{i}^{c_i}$, it first concretizes knowledge into the knowledge of each concept (by converting the original dimensions into the number of concepts through an MLP) and then the comprehensive knowledge state is obtained through a Sigmoid function.
\end{sloppypar}

\begin{sloppypar}
\section{Experiments}

\label{sec:experiments}

Our goal is to answer the following research questions:
\begin{itemize}[leftmargin=*]
\item{\textbf{RQ1}: How does DKT2 perform compared to 18 baselines from 8 different categories under applicable input and comprehensive output settings?}
\item \textbf{RQ2}: How do different input settings for KT models with weak applicability and comprehensiveness and multi-concept prediction of various KT models affect their performance?
\item{\textbf{RQ3}: What are the impacts of the components (\textit{e.g.}, the Rasch embedding and IRT prediction) on DKT2?}
\end{itemize}

\subsection{Experimental Setup}
\subsubsection{Datasets}

We conduct extensive experiments on three of the latest large-scale benchmark datasets from different platforms: Assist17~\cite{feng2009addressing}, EdNet~\cite{choi2020ednet}, and Comp~\cite{hu2023ptadisc}. Details of the datasets are provided in Appendix~\ref{apx: apx-dataset}.

\subsubsection{Baselines}

To comprehensively and systematically evaluate the performance of DKT2 and analyze the impact of input-output settings on KT models, we compare DKT2 with 18 DLKT baselines from 8 categories, as mentioned in Sec.~\ref{sec: related_work}. Detailed descriptions of the aforementioned DLKT baselines can be found in Appendix~\ref{apx: apx-baseline}.

\subsubsection{Implementation}

Similar to CL4KT~\cite{lee2022contrastive}, we employ five-fold cross-validation, with folds divided by students. 10\% of the training set is used for model evaluation and also for the early stopping strategy: if the AUC does not improve within 10 epochs during the 300 epochs, the training will be stopped. The averages across five test folds are reported. We focus on the most recent 100 interactions (history length) for each student, as this latest information is crucial for future predictions. During training, all models are trained using the Adam optimizer~\cite{kingma2017adam} with the following settings: batch size is fixed at 512, learning rate is 0.001, dropout rate is 0.05, and embedding dimension is 64. The seed is set to 12405 to reproduce experimental results. Similar to existing DLKT research, our evaluation metrics include two classification metrics,  Area Under the ROC Curve (AUC) and Accuracy (ACC), and one regression metric, Root Mean Square Error (RMSE). Note that our experimental parameter configuration is consistent with CL4KT.

\subsection{Applicable and Comprehensive Performance Comparison (RQ1)}
Under applicable input and comprehensive output settings, we evaluate three common prediction tasks in KT~\cite{liupykt2022}: \texttt{1)} one-step prediction, \texttt{2)} multi-step prediction, and \texttt{3)} prediction with varying history lengths.

\begin{table*}[t]
\resizebox{\textwidth}{!}{%
\centering
\begin{tabular}{@{}c|c|ccc|ccc|ccc@{}}
\toprule[1.2pt]
\multirow{2}{*}{Category}               & \multirow{2}{*}{Model} & \multicolumn{3}{c|}{Assist17}                                                 & \multicolumn{3}{c|}{EdNet}                                                    & \multicolumn{3}{c}{Comp}                                                     \\ \cmidrule(l){3-5} \cmidrule(l){6-8} \cmidrule(l){9-11} 
                                      &                        & \multicolumn{1}{c}{AUC$\uparrow$} & \multicolumn{1}{c}{ACC$\uparrow$} & \multicolumn{1}{c|}{RMSE$\downarrow$} & \multicolumn{1}{c}{AUC$\uparrow$} & \multicolumn{1}{c}{ACC$\uparrow$} & \multicolumn{1}{c|}{RMSE$\downarrow$} & \multicolumn{1}{c}{AUC$\uparrow$} & \multicolumn{1}{c}{ACC$\uparrow$} & \multicolumn{1}{c}{RMSE$\downarrow$} \\ \midrule
\multirow{3}{*}{Deep sequential}      & DKT\ding{51}                    & \multicolumn{1}{c}{0.6621}    & \multicolumn{1}{c}{0.6370}    & \multicolumn{1}{c|}{0.4731}     & \multicolumn{1}{c}{0.6834}    & \multicolumn{1}{c}{0.6451}    & \multicolumn{1}{c|}{0.4687}     & \multicolumn{1}{c}{0.7585}    & \multicolumn{1}{c}{0.8129}    & \multicolumn{1}{c}{0.3681}     \\
                                      & DKT+\ding{51}                  & 0.6668                     & 0.6415                     & 0.4711                     & 0.6884                     & 0.6483                     & 0.4673                     & 0.7593                     & 0.8129                     & 0.3679                         \\
                                      & DKT-F\ding{51}                 & 0.6633                     & 0.6429                     & 0.4724                     & \underline{0.6917}                     & \underline{0.6503}                     & \underline{0.4668}                     & 0.7615                     & 0.8138                     & 0.3672   \\
                                      \midrule
\multirow{7}{*}{Attention-based}      & SAKT\ding{61}                 & 0.6211                     & 0.6108                     & 0.4828                     & 0.6773                     & 0.6415                     & 0.4708                     & 0.7560                     & 0.8123                     & 0.3690                \\
                                      & AKT\ding{55}                  & 0.6789                     & 0.6464                     & 0.4723                     & 0.6855                     & 0.6440                     & 0.4686                     & 0.7601                     & 0.8119                     & 0.3686     \\
                                      & simpleKT\ding{55}              & 0.6709                     & 0.6441                     & 0.4746                     & 0.6865                     & 0.6444                     & 0.4686                     & 0.7633                     & 0.8135                     & 0.3672          \\
                                      & FoLiBiKT\ding{55}              & 0.6771                     & 0.6444                     & 0.4750                     & 0.6849                     & 0.6432                     & 0.4687                     & 0.7599                     & 0.8120                     & 0.3685      \\
                                      & sparseKT\ding{55}             & 0.6674                     & 0.6424                     & 0.4740                     & 0.6856                     & 0.6430                     & 0.4701                     & \textbf{0.7690}                     & \textbf{0.8178}                     & \textbf{0.3604}           \\
                                      & DTransformer\ding{55}          & 0.6480                     & 0.6305                     & 0.4770                     & 0.6727                     & 0.6355                     & 0.4722                     & 0.7551                     & 0.8106                     & 0.3699         \\
                                      & stableKT\ding{55}          & 0.6781                     & 0.6455                     & 0.4751                     & 0.6841                     & 0.6411                     & 0.4695                     & 0.7591                     & 0.8126                     & 0.3683         \\ \midrule
Mamba-based                           & Mamba4KT\ding{51}              & \underline{0.7001}                     & \underline{0.6555}                     & \underline{0.4701}                     & 0.6667                     & 0.6351                     & 0.4764                     & 0.7575                     & 0.8121                     & 0.3687          \\ \midrule
Graph-based                           & GKT\ding{51}                   & 0.6408                   & 0.6185                   & 0.4802                    & 0.6841                  & 0.6361                   & 0.4724                    & 0.7390                   & 0.8055                   & 0.3766                   \\ \midrule
\multirow{2}{*}{Memory-augmented}     & DKVMN\ding{55}                & 0.6505                     & 0.6308                     & 0.4774                     & 0.6778                     & 0.6410                     & 0.4705                     & 0.7534                     & 0.8113                     & 0.3697            \\
                                      & SKVMN\ding{55}                & 0.6350                     & 0.6184                     & 0.4809                     & 0.6800                     & 0.6427                     & 0.4696                     & 0.7220                     & 0.8040                     & 0.3790      \\ \midrule
Adversarial-based                     & ATKT\ding{51}                  & 0.6453                     & 0.6313                     & 0.4821                     & 0.6780                     & 0.6403                     & 0.4714                     & 0.7560                     & 0.8123                     & 0.3688    \\ \midrule
Contrastive learning-based            & CL4KT\ding{55}                & 0.6540                     & 0.6319                     & 0.4783                     & -                          & -                          & -                          & 0.7645                     & 0.8146                       & 0.3669         \\ \midrule
\multirow{2}{*}{Other representative} & Deep-IRT\ding{55}             & 0.6448                     & 0.6268                     & 0.4814                     & 0.6661                     & 0.6317                     & 0.4769                     & 0.7517                     & 0.8108                     & 0.3703       \\
                                      & AT-DKT\ding{51}                 & 0.6720                     & 0.6433                     & 0.4708                     & 0.6888                     & 0.6494                     & 0.4673                     & 0.7655                     & 0.8141                     & 0.3663            \\ \midrule[1.1pt]
\multirow{1}{*}{{\cellcolor[HTML]{CBCEFB}\textbf{Deep sequential$^\ast$}}} & {\cellcolor[HTML]{CBCEFB}\textbf{DKT2}\ding{51}}                 & {\cellcolor[HTML]{CBCEFB}\textbf{0.7042}}                     & {\cellcolor[HTML]{CBCEFB}\textbf{0.6594}}                     & {\cellcolor[HTML]{CBCEFB}\textbf{0.4630}}                     & {\cellcolor[HTML]{CBCEFB}\textbf{0.6929}}                     & {\cellcolor[HTML]{CBCEFB}\textbf{0.6504}}                     & {\cellcolor[HTML]{CBCEFB}\textbf{0.4660}}                     & {\cellcolor[HTML]{CBCEFB}\underline{0.7679}}                        & {\cellcolor[HTML]{CBCEFB}\underline{0.8165}}                     & {\cellcolor[HTML]{CBCEFB}\underline{0.3652}}   \\
                                      \bottomrule[1.3pt]
\end{tabular}
}
\caption{One-step prediction performance of DKT2 and 18 baselines from different categories. The \textbf{best result} is in bold, the \underline{second best} is underlined. \ding{51} indicates strong applicability and comprehensiveness,
\ding{55} indicates weak applicability and comprehensiveness, \ding{61} indicates strong applicability but weak comprehensiveness. - indicates the model fails to be applied to such a large-scale dataset, resulting in a program crash.}
\vspace{-0.7cm}
\label{tab: one-step}
\end{table*}

\subsubsection{One-step Prediction}
\label{sec: one-step}
KT's one-step prediction can provide immediate feedback for ITS and be used for short-term adjustments of personalized learning paths~\cite{corbett1994knowledge}.
Table~\ref{tab: one-step} shows the one-step prediction performance of DKT2 and 18 baselines from 8 different categories in three large-scale datasets. Overall, in this fair large-scale data competition, our DKT2 has emerged as the final winner by a narrow margin. We observe:
\begin{itemize}[leftmargin=*]
\item{Compared to previous research~\cite{ghosh2020context}, under the input-output settings, attention-based models like AKT still generally outperform deep sequential models like DKT, \textbf{suggesting that attention-based models like AKT may be less affected by these settings.}}
\item{The recently proposed Mamba4KT performs well on Assist17, but underperforms compared to DKT on larger-scale datasets like EdNet and Comp. This may be due to mamba's poorer performance in context learning in large-scale experiments, which is consistent with previous research findings~\cite{waleffe2024empirical}.}
\item{DLKT models based on graph, memory augmentation, adversarial, or contrastive learning do not show significant performance improvements. We believe this is because large-scale data contains more noise and diversity, making it challenging for complex models (\textit{e.g.}, graph-based and memory-augmented models) to effectively extract useful information during training. Moreover, large-scale data usually covers various student learning behaviors and knowledge states, meaning that basic models might already be sufficient for effective knowledge tracing, thus the advantages of adversarial-based and contrastive learning-based models are not pronounced.}
\item{Our proposed DKT2 performs almost the best on all metrics across all datasets. This performance improvement can be attributed to the superiority of DKT2, which includes the exponential activation function in sLSTM that helps improve memory and forgetting processes, and the matrix memory introduced in mLSTM that gives DKT2 advantages in large-scale applications and long sequence processing.}
\end{itemize}

\begin{table*}[t!]
\centering
\resizebox{\textwidth}{!}{%
\begin{tabular}{@{}c|ccc|ccc|ccc|ccc@{}}
\toprule[1.2pt]
\multicolumn{1}{c|}{Step}   & \multicolumn{3}{c|}{5}                                                              & \multicolumn{3}{c|}{10}                                                       & \multicolumn{3}{c|}{15}                                                       & \multicolumn{3}{c}{20} \\ \cmidrule(l){1-1} \cmidrule(l){2-4} \cmidrule(l){5-7} \cmidrule(l){8-10} \cmidrule(l){11-13}
\multicolumn{1}{c|}{Metric} & \multicolumn{1}{c}{AUC$\uparrow$}    & \multicolumn{1}{c}{ACC$\uparrow$}    & \multicolumn{1}{c|}{RMSE$\downarrow$} & \multicolumn{1}{c}{AUC$\uparrow$} & \multicolumn{1}{c}{ACC$\uparrow$} & \multicolumn{1}{c|}{RMSE$\downarrow$} & \multicolumn{1}{c}{AUC$\uparrow$} & \multicolumn{1}{c}{ACC$\uparrow$} & \multicolumn{1}{c|}{RMSE$\downarrow$} & \multicolumn{1}{c}{AUC$\uparrow$} & \multicolumn{1}{c}{ACC$\uparrow$} & \multicolumn{1}{c}{RMSE$\downarrow$} \\ \midrule
DKT                         & \multicolumn{1}{c}{0.6244} & \multicolumn{1}{c}{0.6104} & 0.4831                    & 0.6048                  & 0.5978                  & 0.4868                    & 0.5962                  & 0.5960                  & 0.4874                   & 0.5902 & 0.5918 & 0.4890 \\
\midrule
SAKT                        & 0.6103                     & 0.6010                     & 0.4860                    & 0.6013                  & 0.5966                  & 0.4860                    & 0.5989                  & 0.5983                  & 0.4860                   & 0.5961 & 0.5960 & 0.4869 \\
AKT                         & \underline{0.6486}                     & \underline{0.6285}                     & \textbf{0.4763}                    & \underline{0.6321}                  & \underline{0.6213}                  & \textbf{0.4798}                    & \underline{0.6231}                  & \underline{0.6140}                 & \textbf{0.4819}                   & \underline{0.6189} & \underline{0.6134} & \textbf{0.4827} \\ \midrule
Mamba4KT                    & 0.6222                     & 0.6077                     & 0.4869                    & 0.5909                  & 0.5938                  & 0.4876                    & 0.5875                  & 0.5911                  & 0.4880                   & 0.5858 & 0.5907 & 0.4884 \\ \midrule
DKVMN                       & 0.6205                     & 0.6096                     & 0.4851                    & 0.6008                  & 0.5958                  & 0.4880                    & 0.5905                  & 0.5923                  & 0.4879                   & 0.5830 & 0.5856 & 0.4893 \\ \midrule
ATKT                        & 0.6246                     & 0.6186                     & 0.4831                    & 0.6176                  & 0.6139                  & 0.4847                    & 0.6125                  & 0.6118                  & 0.4855                   & 0.6094 & 0.6090 & 0.4865 \\ \midrule
CL4KT                       & 0.6347                     & 0.6186                     & 0.4832                    & 0.6128                  & 0.6037                  & 0.4882                    & 0.6043                        & 0.5987                    & 0.4896                     & 0.5991 & 0.5971 & 0.4890 \\ \midrule
Deep-IRT                    & 0.6100                     & 0.6020                     & 0.4959                    & 0.5867                  & 0.5834                  & 0.5022                    & 0.5737                  & 0.5713                  & 0.5072                   & 0.5652 & 0.5666 & 0.5049 \\
AT-DKT                      & 0.6424                     & 0.6260                     & \underline{0.4782}                    & 0.6271                  & 0.6154                  & 0.4820                    & 0.6206                  & 0.6115                  & 0.4832                   & 0.6170 & 0.6082 & 0.4849 \\ \midrule[1.1pt]
{\cellcolor[HTML]{CBCEFB}\textbf{DKT2}}                        & {\cellcolor[HTML]{CBCEFB}\textbf{0.6496}}                     & {\cellcolor[HTML]{CBCEFB}\textbf{0.6313}}                     & {\cellcolor[HTML]{CBCEFB}\textbf{0.4763}}                    & {\cellcolor[HTML]{CBCEFB}\textbf{0.6335}}                  & {\cellcolor[HTML]{CBCEFB}\textbf{0.6221}}                  & {\cellcolor[HTML]{CBCEFB}\underline{0.4802}}                    & {\cellcolor[HTML]{CBCEFB}\textbf{0.6246}}                  & {\cellcolor[HTML]{CBCEFB}\textbf{0.6160}}                  & {\cellcolor[HTML]{CBCEFB}\underline{0.4822}}                   & {\cellcolor[HTML]{CBCEFB}\textbf{0.6199}} & {\cellcolor[HTML]{CBCEFB}\textbf{0.6148}} & {\cellcolor[HTML]{CBCEFB}\underline{0.4828}} \\ \bottomrule[1.3pt]
\end{tabular}
}
\caption{Multi-step prediction performance of DKT2 and several representative baselines on Assist17. The results for EdNet and Comp can be found in Appendix~\ref{apx: multi-step-experiment}.}
\vspace{-0.5cm}
\label{tab: multi-step}
\end{table*}

\subsubsection{Multi-step Prediction}
\label{sec: multi-step}

KT's accurate multi-step prediction not only provides valuable feedback for selecting and constructing personalized learning materials, but also assists ITS in flexibly adjusting future curriculum based on student needs~\cite{liu2023simplekt}. Table~\ref{tab: multi-step} and Table~\ref{apx: apx-multi-step} in Appendix~\ref{apx: multi-step-experiment} show the multi-step (step=5, 10, 15, 20) prediction performance of DKT2 and several representative baselines from different categories. The main observations are as follows: (1) As the prediction steps increase, the performance of all models consistently decreases. This is due to error accumulation, meaning that small errors in one-step prediction can accumulate over multiple steps, leading to a decrease in multi-step prediction performance. (2) Compared to one-step prediction, attention-based models perform well in multi-step prediction. This is because the attention mechanism can capture long-distance dependencies, making its advantages more apparent. In contrast, Mamba4KT performs poorly, as mamba-based models are highly dependent on context~\cite{lieber2024jamba} and are more susceptible to error accumulation. (3) Our DKT2 generally outperforms all models in multi-step prediction. We can similarly attribute this to the exponential activation function introduced in sLSTM of DKT2, which can mitigate error accumulation by modifying storage decisions, as it allows the model to update its internal state at each step, while the matrix memory introduced in mLSTM provides support for large-capacity storage space.

\begin{figure*}[t]
\centering
\includegraphics[width=1\linewidth]{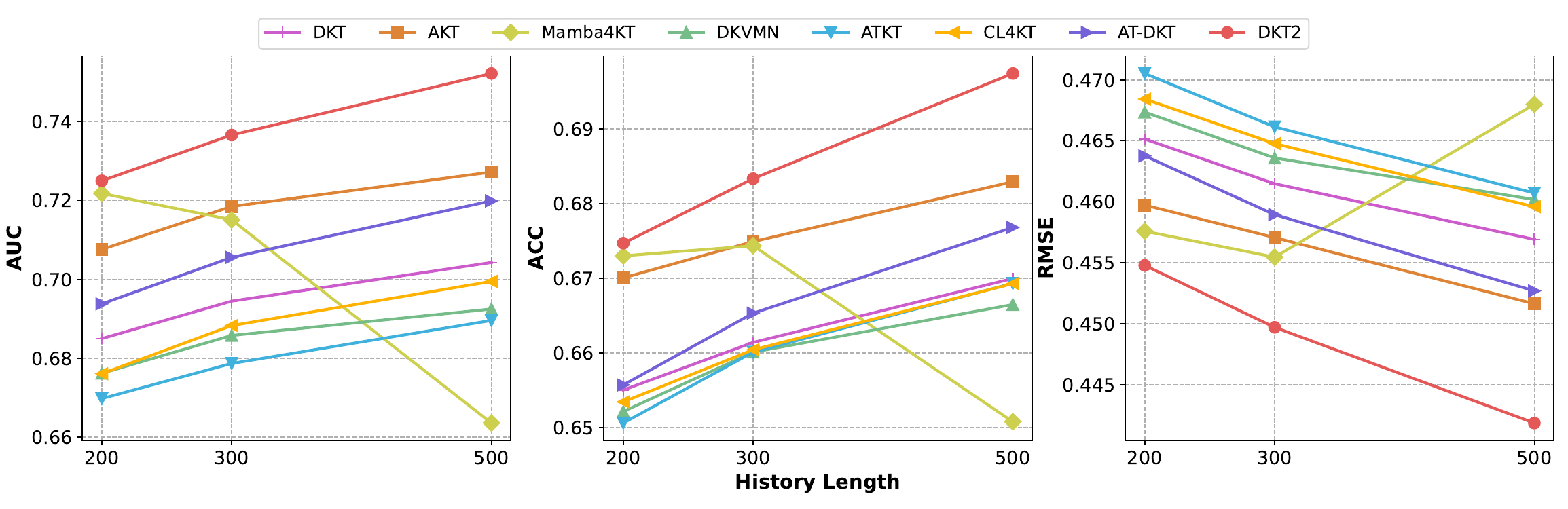}
    \vspace{-0.6cm}
    \caption{The prediction performance of DKT2 and several representative baselines on Assist17 with different history lengths. The results for EdNet and Comp are in Appendix~\ref{apx: multi-step-experiment}.}
    \label{fig: histroy_assist17}
    \vspace{-0.4cm}
\end{figure*}

\subsubsection{Varying-history-length Prediction}
\label{sec: history-length}
Analyzing the impact of different history lengths can help ITS better understand students' knowledge acquisition and forgetting processes, thereby improving teaching strategies. Fig.~\ref{fig: histroy_assist17}, Fig.~\ref{fig: histroy_ednet} and Fig.~\ref{fig: histroy_comp} in Appendix~\ref{apx: varying-length-experiment} show the prediction performance of DKT2 and several representative baselines with different history lengths. From these, we can observe: 1) As the history length increases, the prediction performance of almost all models generally improves, as longer sequences provide more historical information. Surprisingly, Mamba4KT's performance consistently decreases. A possible reason is that mamba-based models are better at capturing local temporal dependencies but may struggle to effectively capture long-distance dependencies within longer sequences. 2) Notably, DKT, using only one LSTM, can maintain a strong ranking position across different history lengths, further encouraging KT researchers to design simple yet effective models~\cite{liu2023simplekt}. 3) Our DKT2 maintains optimal performance across different history lengths, with more significant performance improvements as the history length increases. This is not only due to the increased storage capacity of mLSTM but also related to sLSTM providing a broader output range as the sequence length increases.

\subsection{In-Depth Analysis (RQ2 \& RQ3)}

\begin{table*}[t]
\centering
\resizebox{\textwidth}{!}{%
\begin{tabular}{@{}c|c|c|c|c|c|c|c|c|c|c@{}}
\toprule[1.2pt]
\multicolumn{1}{c|}{Setting} & Metric & AKT & simpleKT & FoLiBiKT & sparseKT & DTransformer & stableKT & DKVMN & CL4KT & Deep-IRT \\
\midrule 
\multicolumn{1}{c}{\cellcolor{gray!10}}
& AUC$\uparrow$ & 0.6554 & 0.6507 & 0.6545 & 0.6405 & 0.5995 & 0.6490 & 0.6228 & 0.5941 & 0.6234 \\
\multicolumn{1}{c}{\cellcolor{gray!10}
$\text{\tiny $\vartriangle$}$} & ACC$\uparrow$ & 0.6154 & 0.6129 & 0.6117 & 0.6120 & 0.5755  & 0.6159 & 0.5899 & 0.5696 & 0.5915 \\
\multicolumn{1}{c}{\cellcolor{gray!10}}
& RMSE$\downarrow$ & 0.4822 & 0.4840 & 0.4835 & 0.4865 & 0.5040 & 0.4837 & 0.4890 & 0.5090 & 0.4892 \\
\cmidrule(l){1-10}
\multicolumn{1}{c}{\cellcolor{gray!10}}
& AUC$\uparrow$ & 0.6505 & 0.6675 & 0.6471 & 0.6574 & 0.5989 & 0.6329 & 0.6203 & 0.6293 & 0.6182 \\
\multicolumn{1}{c}{\cellcolor{gray!10}
$\circ $} & ACC$\uparrow$ & 0.6202 & 0.6240 & 0.6226 & 0.6210 & 0.5625 & 0.6045 & 0.5862 & 0.6016 & 0.5884 \\
\multicolumn{1}{c}{\cellcolor{gray!10}}
& RMSE$\downarrow$ & 0.4853 & 0.4866 & 0.4842 & 0.4836 & 0.5206 & 0.4908 & 0.5010 & 0.5037 & 0.5015 \\
\cmidrule(l){1-10}
\multicolumn{1}{c}{\cellcolor{gray!10}}
& AUC$\uparrow$ & 0.6320 & 0.6508 & 0.6192 & 0.6474 & 0.5994 & 0.6533 & 0.6087 & 0.6195 & 0.6001 \\
\multicolumn{1}{c}{\cellcolor{gray!10}
$\bullet$} & ACC$\uparrow$ & 0.6066 & 0.6153 & 0.5980 & 0.6060 & 0.5770 & 0.6223 & 0.5787 & 0.5933 & 0.5728 \\
\multicolumn{1}{c}{\cellcolor{gray!10}}
& RMSE$\downarrow$ & 0.4881 & 0.4999 & 0.4944 & 0.4944 & 0.4981 & 0.4833 & 0.5074 & 0.5083 & 0.5012 \\
\bottomrule[1.2pt]
\end{tabular}
}
\caption{The prediction performance of KT models with weak applicability and comprehensiveness in the last 5 steps on Assist17 under three different input settings. The $\text{\tiny $\vartriangle$}$ setting represents masking all interaction information (including questions, concepts and responses) for the last 5 steps, the $\circ$ setting represents masking the responses for the last 5 steps, without masking questions and concepts, and the $\bullet$ setting represents no masking, \textit{i.e.}, predicting the responses under the regular setting. The results for EdNet and Comp can be found in Appendix~\ref{apx: multi-step-experiment}.}
\label{tab: settings}
\vspace{-0.6cm}
\end{table*}

\subsubsection{Analysis of Different Input Settings}

We analyze three different input settings for the KT models with weak applicability and comprehensiveness. In Table~\ref{tab: settings} and Table~\ref{apx: apx-settings} in Appendix~\ref{apx: input-setting-experiment}, we present the prediction performance of these models in the last 5 steps. From these, we have the following findings: (i) The performance differences among these three settings are more pronounced on EdNet and Comp, as larger-scale data can provide richer information for more accurate prediction. (ii) The models under guessing $\text{\tiny $\vartriangle$}$ setting seems to perform well on Assist17, which may be because the models remember the answer bias~\cite{cui2023we} and make predictions directly, while the models under the $\circ$ and $\bullet$ settings achieve comparable performance, indicating that \textbf{the applicable $\circ$ setting does not significantly reduce the model's performance.} This confirms the hypothesis proposed in the Sec.~\ref{sec: one-step} (One-step Prediction).

\begin{table*}[t]
\centering
\resizebox{\textwidth}{!}{%
\begin{tabular}{@{}c|ccc|ccc|ccc@{}}
\toprule[1.2pt]
Dataset  & \multicolumn{3}{c|}{Assist17}                                                                       & \multicolumn{3}{c|}{EdNet}                                                                          & \multicolumn{3}{c}{Comp}                                                                            \\ \midrule
Metric   & AUC$\uparrow$                             & ACC$\uparrow$                             & RMSE$\downarrow$                            & AUC$\uparrow$                             & ACC$\uparrow$                             & RMSE$\downarrow$                            & AUC$\uparrow$                             & ACC$\uparrow$                             & RMSE$\downarrow$                            \\ \midrule
DKT      & 0.5841                          & 0.5787                          & 0.4913                          & 0.6600                          & 0.6225                          & 0.4775                          & 0.7091                          & 0.8045                          & 0.3806                          \\ \midrule
SAKT     & 0.5596                          & 0.5534                          & 0.5048                          & 0.6546                          & 0.6198                          & 0.4788                          & 0.6994                          & 0.8037                          & 0.3824                          \\
AKT      & \underline{0.6185} & \underline{0.6040} & 0.4862                          & \underline{0.6649} & 0.6241                          & \textbf{0.4765}                 & 0.7054                          & 0.8039                          & 0.3815                          \\ \midrule
Mamba4KT & 0.5660                          & 0.5660                          & 0.4956                          & 0.6531                          & 0.6192                          & 0.4796                          & 0.7054                          & 0.8043                          & 0.3813                          \\ \midrule
DKVMN    & 0.5730                          & 0.5701                          & 0.4964                          & 0.6572                          & 0.6205                          & 0.4781                          & 0.7050                          & 0.8034                          & 0.3817                          \\ \midrule
ATKT     & \underline{0.6205} & \underline{0.6077} & \underline{0.4836} & 0.6639                          & 0.6229                          & 0.4768                          & \underline{0.7111} & \underline{0.8049} & \underline{0.3802} \\ \midrule
CL4KT    & 0.5892                          & 0.5911                          & 0.4890                          & -                               & -                               & -                               & 0.7044                          & 0.8032                          & 0.3820                          \\ \midrule
Deep-IRT & \textbf{0.6445}                 & \textbf{0.6263}                 & \textbf{0.4808}                 & \textbf{0.6664}                 & \textbf{0.6321}                 & 0.4767                          & \textbf{0.7515}                 & \textbf{0.8107}                 & \textbf{0.3704}                 \\
AT-DKT   & 0.6087                          & 0.5962                          & 0.4882                          & 0.6644                          & \underline{0.6245} & \underline{0.4766} & 0.7093                          & 0.8046                          & 0.3805                          \\ \midrule
{\cellcolor[HTML]{CBCEFB}DKT2}     & {\cellcolor[HTML]{CBCEFB}0.6174}                          & {\cellcolor[HTML]{CBCEFB}0.6041}                          & {\cellcolor[HTML]{CBCEFB}0.4872}                          & {\cellcolor[HTML]{CBCEFB}0.6646}                          & {\cellcolor[HTML]{CBCEFB}0.6243}                          & {\cellcolor[HTML]{CBCEFB}0.4768}                          & {\cellcolor[HTML]{CBCEFB}0.7064}                          & {\cellcolor[HTML]{CBCEFB}0.8047}                          & {\cellcolor[HTML]{CBCEFB}0.3810}                          \\ \bottomrule
\end{tabular}
}
\caption{Multi-concept prediction performance of DKT2 and several representative baselines.}
\label{apx: apx-multi-concept}
\vspace{-0.4cm}
\end{table*}

\subsubsection{Multi-concept Prediction}
\label{sec: output}

Comprehensive KT can be used for multi-concept prediction. Multi-concept prediction can provide a more comprehensive learning assessment, explore relationships between concepts, and create precise personalized learning plans for students. Due to the lack of datasets for multi-concept prediction (to our knowledge, existing datasets do not include students' proficiency scores for all concepts at different learning stages), our experiments are conducted under a weak assumption: the change in a student's knowledge state is a gradual process and is unlikely to experience sudden shifts over the long term. In our experiments, we use the knowledge state at the intermediate time step to predict subsequent questions. Table~\ref{apx: apx-multi-concept} shows the multi-concept prediction performance of DKT2 and several representative baselines. From this, we discover an unexpected phenomenon: Deep-IRT and ATKT, which are generally not advantageous in previous performance comparisons, achieve impressive results, while our DKT2 can only rank in the top four. These results might make us question the validity of the weak assumption, but the empirical evidence of the almost consistent performance rankings of Deep-IRT and ATKT across the three datasets dispels our doubts. This interesting phenomenon makes us ponder: is it necessary to excessively pursue prediction accuracy while neglecting the assessment of multiple concepts in practice? We will explore this important topic in depth in future KT research.

\subsubsection{Ablation Study}
\label{sec: ablation}

\begin{figure}[t]
\includegraphics[width=\linewidth]{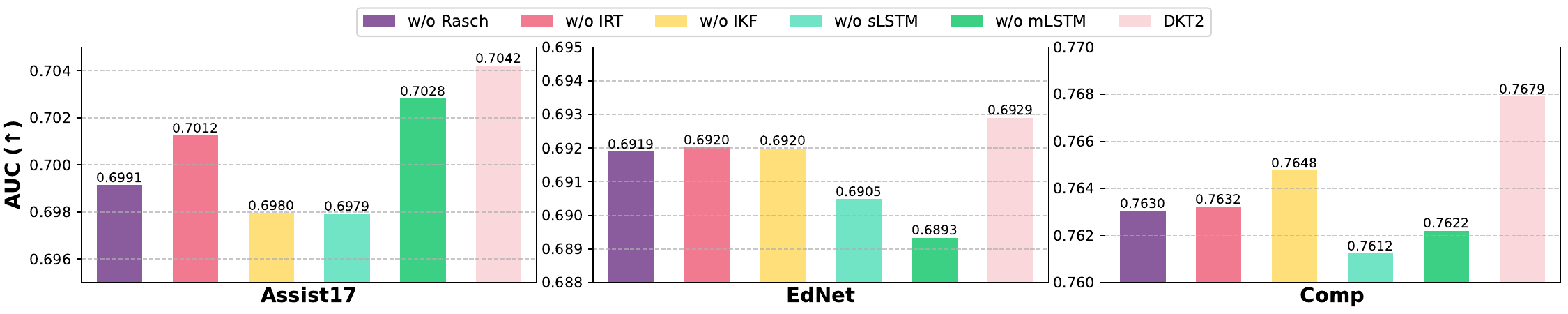}
    \vspace{-0.7cm}
    \caption{Ablation study on AUC.}
    \label{fig: ablation_auc}
    \vspace{-0.4cm}
\end{figure}

Fig.~\ref{fig: ablation_auc} and Fig.~\ref{fig: ablation_acc_rmse} in Appendix~\ref{apx: ablation} illustrate the impact of different components on DKT2. ``w/o. Rasch" indicates the removal of Rasch embedding from DKT2 (setting $d_{q_t}$ to $0$ in Eq.~\ref{eq: embedding}), ``w/o. IRT" represents the removal of the IRT module, ``w/o. IKF" means DKT2 ignores the integrated knowledge fusion, while ``w/o. sLSTM" and ``w/o. mLSTM" denote the removal of sLSTM block and mLSTM block, respectively. The results show that DKT2 achieves the highest AUC scores across all datasets compared to other variants, demonstrating the importance of each component on DKT2. Notably, ``w/o. mLSTM" generally outperforms DKT2 on ACC and RMSE scores on Assist17, which is due to mLSTM's inability to demonstrate significant advantages in small-scale data, as evidenced by its poorer performance on larger datasets, EdNet and Comp.

\end{sloppypar}

\section{Conclusion}
\label{sec:conclusion}

This paper introduces DKT2, an applicable and comprehensive DLKT model that addresses key limitations of deep sequential models like DKT. By leveraging xLSTM, the Rasch model, and Item Response Theory (IRT), DKT2 effectively balances predictive performance with practical applicability. Our extensive experiments across three large-scale datasets demonstrate DKT2's superiority over 18 baseline models in various prediction tasks, highlighting its robustness and potential for real-world educational applications.

\section{Limitations}
\label{sec: limitation}

Our work represents an attempt to apply xLSTM in the KT domain on large-scale data with fair input and output settings. In our experiments, we observe that as the number of students increases, DKT2 gradually demonstrates a performance advantage that widens the gap with other DLKT models. Additionally, in our multi-concept prediction experiments, we find that Deep-IRT exhibits a leading, dataset-independent advantage, the reasons for which give us pause for reflection. Therefore, our future research directions include: 1) further exploration of deeper knowledge tracing methodologies based on xLSTM, particularly in the context of ultra-large-scale data, and 2) enhancing multi-concept predictive analysis by collecting and analyzing students' proficiency scores across different concepts at various learning stages.

\section*{Acknowledgments}
This research was supported by grants from the "Pioneer" and "Leading Goose" R\&D Program of Zhejiang (2025C02022), National Natural Science Foundation of China (No.62307032), and Shanghai Rising-Star Program (23QA1409000).

\bibliographystyle{splncs04}
\bibliography{reference}

\clearpage
\appendix

\begin{table*}[htbp]
\setlength{\tabcolsep}{13mm}
\resizebox{\textwidth}{!}{%
\centering
\begin{tabular}{@{}cccc@{}}
\toprule[1.5pt]
\rowcolor{gray!10} \multicolumn{1}{c}{\textbf{Conference/Journal}}     & \textbf{Model}        & \textbf{Applicability} & \multicolumn{1}{c}{\textbf{Comprehensiveness}} \\ \midrule \midrule
\multirow{4}{*}{AAAI}  & \textbf{KTM}~\cite{vie2019knowledge}          & \Checkmark          & \XSolidBrush             \\
                        & \textbf{IKT}~\cite{minn2022interpretable}          & \Checkmark          & \XSolidBrush             \\
                       & 
                       \textbf{QIKT}~\cite{chen2023improving}         & \Checkmark          & \Checkmark             \\
                       & 
                       \colorbox{gray!15}{\textbf{DAKTN}~\cite{wang2023deep}}         & \colorbox{gray!15}{\Checkmark}          & \colorbox{gray!15}{\Checkmark}             \\ \midrule
\multirow{9}{*}{CIKM}  & \textbf{MF-DAKT}~\cite{zhang2021multi}          & \Checkmark          & \XSolidBrush             \\
& \textbf{LFBKT}~\cite{chen2022knowledge}          & \Checkmark          & \XSolidBrush             \\
                        & \textbf{RKT}~\cite{pandey2020rkt}          & \XSolidBrush          & \XSolidBrush             \\
                       & \textbf{FoLiBiKT}~\cite{im2023forgetting}     & \XSolidBrush          & \XSolidBrush             \\
                       & \textbf{CPKT}~\cite{wang2023continuous}     & \Checkmark          & \XSolidBrush             \\
                       & \colorbox{gray!15}{\textbf{SFKT}~\cite{zhang2023no}}     & \colorbox{gray!15}{\Checkmark}          & \colorbox{gray!15}{\XSolidBrush}             \\
                       & \colorbox{gray!15}{\textbf{CMKT}~\cite{zhang2023counterfactual}}     & \colorbox{gray!15}{\Checkmark}          & \colorbox{gray!15}{\XSolidBrush}             \\
                       & \textbf{Sinkt}~\cite{fu2024sinkt}     & \Checkmark          & \XSolidBrush             \\
                       & \colorbox{gray!15}{\textbf{LOKT}~\cite{guo2024mitigating}}     & \colorbox{gray!15}{\Checkmark}          & \colorbox{gray!15}{\XSolidBrush}             \\ \midrule
\multirow{1}{*}{COLING}  & \colorbox{gray!15}{\textbf{KVFKT}~\cite{guan2025kvfkt}}          & \colorbox{gray!15}{\Checkmark}          & \colorbox{gray!15}{\XSolidBrush}             \\ \midrule
\multirow{3}{*}{ICDM}  & \textbf{DKT-DSC}~\cite{minn2018deep}      & \Checkmark          & \Checkmark             \\
                       & \textbf{SKT}~\cite{tong2020structure}          & \Checkmark          & \XSolidBrush             \\
                       & \textbf{CAKT}~\cite{yang2022deep}         & \Checkmark          & \XSolidBrush             \\ \midrule
\multirow{2}{*}{ICLR}                   & \textbf{simpleKT}~\cite{liu2023simplekt}     & \XSolidBrush          & \XSolidBrush             \\
& \textbf{PSI-KT}~\cite{zhou2024predictive}     & \Checkmark          & \XSolidBrush             \\ \midrule
\multirow{1}{*}{IJCAI}      
& \textbf{stableKT}~\cite{li2024enhancing}     & \XSolidBrush          & \XSolidBrush             \\ \midrule
\multirow{5}{*}{KDD}   & \textbf{AKT}~\cite{ghosh2020context}          & \XSolidBrush          & \XSolidBrush             \\
                       & \textbf{LPKT}~\cite{shen2021learning}         & \XSolidBrush          & \XSolidBrush             \\
                       & \textbf{LBKT}~\cite{xu2023learning}         & \XSolidBrush          & \XSolidBrush             \\ 
                       & \textbf{DyGKT}~\cite{cheng2024dygkt}         & \Checkmark          & \XSolidBrush             \\ 
                       & \textbf{GRKT}~\cite{cui2024leveraging}         & \Checkmark          & \XSolidBrush             \\ \midrule
\multirow{4}{*}{MM}    & \textbf{ATKT}~\cite{guo2021enhancing}         & \Checkmark          & \Checkmark             \\
                       & \textbf{ABQR}~\cite{sun2023adversarial}         & \Checkmark          & \XSolidBrush             \\
                       & \textbf{PSKT}~\cite{huang2024remembering}         & \Checkmark          & \Checkmark             \\
                       & \textbf{ReKT}~\cite{shen2024revisiting}         & \Checkmark          & \XSolidBrush             \\\midrule
NIPS                   & \textbf{DKT}~\cite{NIPS2015_bac9162b}          & \Checkmark          & \Checkmark             \\ \midrule
\multirow{3}{*}{PKDD}  & \textbf{GIKT}~\cite{yang2021gikt}         & \Checkmark          & \XSolidBrush             \\
                       & \textbf{GMKT}~\cite{zhao2023graph}         & \Checkmark          & \XSolidBrush             \\
                       & \colorbox{gray!15}{\textbf{CCKT}~\cite{zheng2024co}}         & \colorbox{gray!15}{\Checkmark}          & \colorbox{gray!15}{\XSolidBrush}             \\\midrule
\multirow{6}{*}{SIGIR} & \textbf{SKVMN}~\cite{abdelrahman2019knowledge}        & \XSolidBrush          & \XSolidBrush             \\
                    & \colorbox{gray!15}{\textbf{HGKT}~\cite{tong2022introducing}}        & \colorbox{gray!15}{\Checkmark}          & \colorbox{gray!15}{\XSolidBrush}             \\
                       & \textbf{CKT}~\cite{shen2020convolutional}          & \Checkmark          & \XSolidBrush             \\
                       & \textbf{IEKT}~\cite{long2021tracing}         & \XSolidBrush          & \XSolidBrush             \\
                       & \textbf{DIMKT}~\cite{shen2022assessing}     & \Checkmark          & \XSolidBrush             \\
                       & \textbf{sparseKT}~\cite{huang2023towards}     & \XSolidBrush          & \XSolidBrush             \\ \midrule
\multirow{4}{*}{TKDE}                   & \textbf{EKT}~\cite{liu2019ekt}       & \Checkmark          & \XSolidBrush             \\
& \textbf{DGMN}~\cite{abdelrahman2022deep}       & \Checkmark          & \Checkmark             \\
& \textbf{LPKT-S}~\cite{shen2022monitoring}       & \XSolidBrush          & \XSolidBrush             \\ 
& \colorbox{gray!15}{\textbf{XKT}~\cite{huang2024xkt}}       & \colorbox{gray!15}{\Checkmark}          & \colorbox{gray!15}{\XSolidBrush}             \\ \midrule
\multirow{5}{*}{TOIS}                   & \colorbox{gray!15}{\textbf{MRT-KT}~\cite{cui2023fine}}       & \colorbox{gray!15}{\XSolidBrush}          & \colorbox{gray!15}{\XSolidBrush}             \\
& \textbf{DGEKT}~\cite{cui2024dgekt}       & \Checkmark          & \Checkmark             \\ 
& \textbf{MAN}~\cite{he2023man}       & \XSolidBrush          & \XSolidBrush             \\
& \textbf{FDKT}~\cite{liu2024fdkt}       & \Checkmark          & \XSolidBrush             \\
& \colorbox{gray!15}{\textbf{ELAKT}~\cite{pu2024elakt}}       & \colorbox{gray!15}{\Checkmark}          & \colorbox{gray!15}{\Checkmark}             \\ \midrule
\multirow{3}{*}{WSDM}  & \textbf{HawkesKT}~\cite{wang2021temporal}     & \XSolidBrush          & \XSolidBrush             \\
& \colorbox{gray!15}{\textbf{AdaptKT}~\cite{cheng2022adaptkt}}         & \colorbox{gray!15}{\Checkmark}          & \colorbox{gray!15}{\XSolidBrush}             \\
                       & \textbf{CoKT}~\cite{long2022improving}         & \Checkmark          & \XSolidBrush             \\ \midrule
\multirow{9}{*}{WWW}   & \textbf{DKVMN}~\cite{zhang2017dynamic}        & \XSolidBrush          & \XSolidBrush             \\
                       & \textbf{DKT-F}~\cite{nagatani2019augmenting}        & \Checkmark          & \Checkmark             \\
                       & \textbf{CL4KT}~\cite{lee2022contrastive}        & \XSolidBrush          & \XSolidBrush             \\
                       & \textbf{DTransformer}~\cite{yin2023tracing} & \XSolidBrush          & \XSolidBrush             \\
                       & \textbf{AT-DKT}~\cite{liu2023enhancing}       & \Checkmark          & \Checkmark             \\
                       & \textbf{MIKT}~\cite{sun2024interpretable}         & \XSolidBrush          & \XSolidBrush             \\
                       & \colorbox{gray!15}{\textbf{QDCKT}~\cite{liu2024question}}         & \colorbox{gray!15}{\Checkmark}          & \colorbox{gray!15}{\XSolidBrush}             \\
                       & \colorbox{gray!15}{\textbf{HD-KT}~\cite{ma2024hd}}         & \colorbox{gray!15}{\Checkmark}          & \colorbox{gray!15}{\XSolidBrush}             \\
                       & \textbf{DisKT}~\cite{zhou2025disentangled}         & \XSolidBrush          & \XSolidBrush             \\\bottomrule[1.5pt]
\end{tabular}%
}
\caption{Summary of the applicability and comprehensiveness of DLKT models in top AI/ML conferences/journals from 2015-2025. \Checkmark and \XSolidBrush indicate strong and weak applicability and comprehensiveness, respectively. The \colorbox{gray!15}{gray background} indicates that the code is not open-source, and its applicability and comprehensiveness are inferred from the method section of the paper.}
\label{tab: summary}
\end{table*}

\section{Appendix}
\setlength{\abovedisplayskip}{3pt}
\setlength{\belowdisplayskip}{3pt}

Due to space limitations, the main text cannot include all details. Here, we have supplemented the details mentioned in the main text, including:
\begin{itemize}[leftmargin=*]
    \item{Summary of DLKT Models from 2015-2025 in Terms of Applicability and Comprehensiveness (\ref{apx: apx-summary})}
    \item{Detailed Introduction to LSTM (\ref{apx: apx-lstm})}
    \item{Detailed Introduction to xLSTM (\ref{apx: apx-xlstm})}
    \item{Dataset Description and Processing Methods (\ref{apx: apx-dataset})}
    \item{Baseline Description (\ref{apx: apx-baseline})}
    \item{Additional Experimental Results (\ref{apx: experiment})}
\end{itemize}

\subsection{Summary of DLKT Models from 2015-2025 in Terms of Applicability and Comprehensiveness}
\label{apx: apx-summary}

Table~\ref{tab: summary} summarizes the DLKT models in terms of applicability and comprehensiveness in top AI/ML conferences/journals from 2015-2025.

\subsection{Detailed Introduction to LSTM}
\label{apx: apx-lstm}

\begin{figure}
    \centering
    \includegraphics[width=0.7\linewidth]{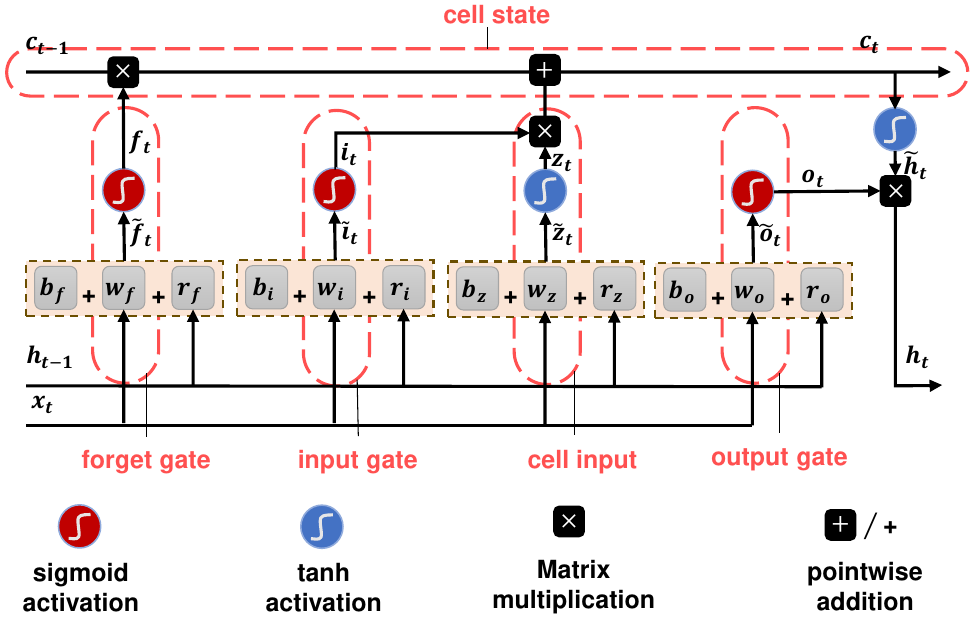}
    \caption{Architecture of LSTM.}
    \label{fig: lstm}
    \vspace{-0.3cm}
\end{figure}

Long Short-Term Memory (LSTM)~\cite{hochreiter1997long} overcomes the short-term memory limitations of Recurrent Neural Networks (RNN)~\cite{elman1990finding} caused by the vanishing gradient\footnote{The vanishing gradient refers to the phenomenon where, during model training, as time step increases, the gradient is continuously multiplied by the weight matrix during backpropagation, potentially causing it to shrink rapidly towards zero, resulting in very slow weight updates in the network.}~\cite{hochreiter1991untersuchungen,hochreiter2001gradient} by introducing cell state and gating mechanisms into the network. Fig.~\ref{fig: lstm} shows the architecture of LSTM at time step $t$. The core concepts of LSTM include cell state and various gate structures. The cell state acts as a pathway for transmitting relevant information, allowing information to be passed along the sequence chain, which can be viewed as the network's memory. Theoretically, during sequence processing, the cell state can continuously carry relevant information. Thus, information obtained at earlier time steps can be transmitted to cells at later time steps, which helps mitigate the impact of short-term memory. Additionally, LSTM addresses the short-term memory issue of RNNs by introducing internal gating mechanisms (\textit{i.e.}, forget gate~\cite{gers2000learning}, input gate, and output gate) to regulate information flow. Specifically, LSTM uses the Tanh activation function (with output values always in the range (-1, 1)) to help regulate the neural network output and employs the Sigmoid activation function in its gate structures. The Sigmoid function is similar to the Tanh function, but its output range is (0, 1), which aids in updating or forgetting data, as any number multiplied by 0 becomes 0 (this information is forgotten), and any number multiplied by 1 remains unchanged (this information is fully preserved). This allows the network to understand which data is unimportant and should be forgotten, and which data is important and should be preserved. The cell state update rule (\textit{i.e.}, the constant error carousel~\cite{hochreiter1996lstm}) for LSTM at time step $t$ is:

\begin{equation}
\begin{split}
    f_t &= \sigma(\tilde{f}_t),\quad \tilde{f}_t = w_f^{\top}\,x_t+r_f\,h_{t-1}+b_f, \\
    i_t &= \sigma(\tilde{i}_t),\quad \tilde{i}_t = w_i^{\top}\,x_t+r_i\,h_{t-1}+b_i, \\
    z_t &= \varphi(\tilde{z}_t),\quad \tilde{z}_t = w_z^{\top}\,x_t+r_z\,h_{t-1}+b_z, \\
    c_t &= f_t\, c_{t-1} + i_t \, z_t, \\
    o_t &= \sigma(\tilde{o}_t),\quad \tilde{o}_t = w_o^{\top}\,x_t+r_o\,h_{t-1}+b_o, \\
    h_t &= o_t \, \tilde{h}_t, \quad \tilde{h}_t = \varphi(c_t), 
\end{split}
\end{equation}

where the weight vectors $w_f, w_i, w_z$, and $w_o$ correspond to the input weights between the input $x_t$ and the forget gate, input gate, cell input, and output gate, respectively. The weights $r_f, r_i, r_z$, and $r_o$ correspond to the recurrent weights between the hidden state $h_{t-1}$ and the forget gate, input gate, cell input, and output gate, respectively. $b_f, b_i, b_z$, and $b_o$ are the corresponding bias terms. $\varphi(\cdot)$ is the activation function for the cell input or hidden state (\textit{e.g.}, Tanh), and $\sigma(\cdot)$ is the Sigmoid activation function, \textit{i.e.}, $\sigma(x)=\frac{1}{1+exp(-x)}$.

In summary, the forget gate in LSTM determines which relevant information from previous time steps should be preserved, the input gate decides which important information from the current input should be added, and the output gate determines the next hidden state. Previous work~\cite{greff2016lstm} has shown that each gate structure is crucial. Recently, LSTM has been revisited and greatly improved, with the revised LSTM known as xLSTM~\cite{beck2024xlstm}. xLSTM enhances the traditional LSTM structure, aiming to improve LSTM's performance and scalability with large-scale data. Subsequently, a series of studies on xLSTM have been applied to various fields such as computer vision~\cite{alkin2024vision,zhu2024seg,dutta2024vision} and time series~\cite{alharthi2024xlstmtime}.

\subsection{Detailed Introduction to xLSTM}
\label{apx: apx-xlstm}
\subsubsection{Stabilized Long Short-Term Memory}

To enable LSTM to revise storage decisions, sLSTM introduces an exponential activation function along with normalizer state and stabilization. Unlike the Sigmoid activation function (\textit{i.e.}, S-shaped function) mentioned in Appendix~\ref{apx: apx-lstm}, where it becomes very challenging for the model to decide what to forget or retain as input values get higher, sLSTM uses an exponential function instead, providing a broader output range, indicating that sLSTM can better revise storage decisions. However, after introducing the exponential function, output values tend to surge as input values increase and do not naturally normalize outputs between 0 and 1 as the Sigmoid function does. Therefore, sLSTM introduces normalizer state, which is a function of the forget gate and input gate, to normalize the hidden state. The update rule for the sLSTM cell state at time step $t$ is:
\setlength{\arraycolsep}{2pt}
\begin{equation}
\begin{split}
    f_t &= \sigma(\tilde{f}_t)\, \text{OR}\, \text{exp}(\tilde{f}_t),\\
    \tilde{f}_t &= w_f^{\top}\,x_t+r_f\,h_{t-1}+b_f, \\
    i_t &= \text{exp}(\tilde{i}_t),\quad \tilde{i}_t = w_i^{\top}\,x_t+r_i\,h_{t-1}+b_i, \\
    z_t &= \varphi(\tilde{z}_t),\quad \tilde{z}_t = w_z^{\top}\,x_t+r_z\,h_{t-1}+b_z, \\
    c_t &= f_t\, c_{t-1} + i_t \, z_t, \\
    n_t &= f_t\, n_{t-1} + i_t, \\
    o_t &= \sigma(\tilde{o}_t),\quad \tilde{o}_t = w_o^{\top}\,x_t+r_o\,h_{t-1}+b_o, \\
    h_t &= o_t \, \tilde{h}_t, \quad \tilde{h}_t = c_t/n_t, 
\end{split}
\end{equation}
where the weight vectors $w_f, w_i, w_z,$ and $w_o$ correspond to the input weights between the input $x_t$ and the forget gate, input gate, cell input, and output gate, respectively. The weights $r_f, r_i, r_z,$ and $r_o$ correspond to the recurrent weights between the hidden state $h_{t-1}$ and the forget gate, input gate, cell input, and output gate, respectively. $b_f, b_i, b_z,$ and $b_o$ are the corresponding bias terms. $\varphi$ is the activation function for the cell input or hidden state (\textit{e.g.}, Tanh), $\sigma$ is the Sigmoid activation function, and \text{exp} is the exponential activation function.

Moreover, since the exponential activation function can easily cause overflow for large values, to prevent the exponential function from disrupting the forget gate and input gate, sLSTM uses an additional state $m_t$~\cite{milakov2018online}, which appears in logarithmic form, to counteract the effect of the exponential function and introduce stability:
\vspace{-2pt}
\begin{equation}
\begin{split}
    &m_t = \text{max}(\text{log}(f_t) + m_{t-1}, \text{log}(i_t)), \\
    &i_t^\prime = \text{exp}(\text{log}(i_t)-m_t)=\text{exp}(\tilde{t}_t - m_t), \\
    &f_t^\prime = \text{exp}(\text{log}(f_t)+m_{t-1}-m_t), 
\end{split}
\label{eq: stabilizer}
\end{equation}
\vspace{-2pt}

\subsubsection{Matrix Long Short-Term Memory}

To enhance LSTM's memory ability to capture more complex data relationships and patterns, mLSTM introduces a matrix $C\in \mathbb{R}^{d\times d}$ to replace the scalar cell state $c\in \mathbb{R}$. Additionally, since LSTM is designed to process sequential data, which means it needs to process the output of the previous input in the sequence to handle the current input, this hinders parallelization and is the main culprit leading to the Transformer era. Therefore, mLSTM abandons this design concept. Specifically, mLSTM adopts the setting of Bidirectional Associative Memories (BAMs)~\cite{kohonen1972correlation,anderson1977distinctive}: at time step $t$, mLSTM stores a pair of vectors, key $k_t\in\mathbb{R}^d$ and value $v_t\in\mathbb{R}^d$. At time step $t+\tau$, the value $v_t$ is retrieved through a query vector $q_{t+\tau}\in \mathbb{R}^d$. mLSTM uses a covariance update rule ($C_t = C_{t-1} + v_t\, k_t^\top$) to store the key-value pair. The covariance update rule is equivalent to the Fast Weight Programmer~\cite{schmidhuber1992learning}. Later, a new variant has emerged~\cite{ba2016using}: a constant decay rate multiplied by $C_{t-1}$ and a constant learning rate multiplied by $v_t\, k_t^\top$. Similarly, in mLSTM, the forget gate corresponds to the decay rate, while the input gate corresponds to the learning rate. Furthermore, since the dot product between the query input and the normalizer state may approach zero, mLSTM uses the absolute value of the dot product and sets a lower bound to a threshold (\textit{e.g.}, 1). The cell state update rule for mLSTM is:
\setlength{\arraycolsep}{2pt}
\begin{equation}
\begin{split}
    f_t &= \sigma(\tilde{f}_t)\, \text{OR}\, \text{exp}(\tilde{f}_t),\quad \tilde{f}_t = w_f^{\top}\,x_t+b_f, \\
    i_t &= \text{exp}(\tilde{i}_t),\quad \tilde{i}_t = w_i^{\top}\,x_t+b_i, \\
    k_t &= \frac{1}{\sqrt{d}}W_k\, x_t+b_k, \\
    v_t &= W_v\, x_t+b_v, \\
    q_t &= W_q\, x_t+b_q, \\
    C_t &= f_t\, C_{t-1} + i_t \, v_t\, k_t^\top, \\
    n_t &= f_t\, n_{t-1} + i_t\, k_t, \\
    o_t &= \sigma(\tilde{o}_t),\quad \tilde{o}_t = W_o^{\top}\,x_t+b_o, \\
    h_t &= o_t \odot \tilde{h}_t, \quad \tilde{h}_t = C_t\, q_t/\text{max}\{\left |n_t^\top\, q_t \right |, 1\},  
\end{split}
\end{equation}

Similarly, to stabilize the exponential function in mLSTM, mLSTM employs the same stabilization technique as sLSTM (see Eq.~\ref{eq: stabilizer}). The design of mLSTM supports highly parallelized processing, which not only improves computational efficiency but also allows the model to scale better to large datasets.

In addition, xLSTM introduces residual networks~\cite{he2016deep} to stack sLSTM or mLSTM, enabling xLSTM to effectively process complex sequential data while improving the training stability of the model in deep networks.

\subsection{Dataset Description and Processing Methods}
\label{apx: apx-dataset}
We provide a detailed description of the datasets used in our experiments and the methods employed for processing them.

We conduct extensive experiments on three of the latest large-scale benchmark datasets from different platforms: (i) Assist17\footnote{\url{https://sites.google.com/view/assistmentsdatamining/dataset?authuser=0}} is the latest subset of the ASSISTments dataset released by Worcester Polytechnic Institute. ASSISTments is an online tutoring system that provides mathematics instruction and access services for students, widely used in mathematics courses for 4th to 12th-grade students in the United States. A key feature of ASSISTments is providing students with immediate feedback, allowing them to know whether their answers are correct after responding to questions. (ii) EdNet\footnote{\url{https://github.com/riiid/ednet}} is a substantial educational dataset collected by Santa, a multi-platform artificial intelligence tutoring service. Collected over two years, this dataset encompasses a wide range of student-system interactions across Android, iOS, and web platforms in Korea. It contains over 130 million learning interactions from approximately 780,000 students, making it one of the largest publicly available interactive education system datasets. The dataset is notable for its scale and hierarchical structure, offering rich insights into student activities and learning patterns. To ensure computational efficiency, we randomly selected 20,000 students from EdNet, similar to previous studies~\cite{liu2020improving,lee2022contrastive,cui2023we}. (iii) Comp\footnote{\url{https://github.com/wahr0411/PTADisc}}, which is part of PTADisc, is specifically selected for KT tasks in computational thinking courses. PTADisc originates from PTA, an online programming teaching assistant system developed by PTA Educational Technology Co., Ltd. for universities and society, based on students. PTADisc is currently the largest dataset in the field of personalized learning, which also includes different courses of varying data scales, providing options for various types of learning.

Following the data preprocessing method in CL4KT~\cite{lee2022contrastive}, we exclude students with fewer than five interactions and all interactions involving unnamed concepts. Since a single question may involve multiple concepts, we convert the unique concept combinations within a single question into a new concept. The statistics after processing are shown in Table~\ref{apx: statistics}.

\begin{table}[]
\centering
\vspace{-0.4cm}
\begin{tabular}{@{}cccccc@{}}
\toprule
\rowcolor{gray!10} \textbf{Datasets} & \textbf{\#students} & \textbf{\#questions} & \textbf{\#concepts} & \textbf{\#interactions} \\ \midrule
\midrule
Assist17 & 1,708      & 3,162       & 411         & 934,638        \\
EdNet    & 20,000     & 12,215      & 1,781       & 2,709,132      \\
Comp     & 45,180     & 8,392       & 472         & 6,072,632      \\ \bottomrule
\end{tabular}%
\caption{Statistics of three datasets after processing.}
\vspace{-0.8cm}
\label{apx: statistics}
\end{table}

\begin{table*}[]
\resizebox{\textwidth}{!}{%
\centering
\begin{tabular}{@{}c|c|ccc|ccc|ccc|ccc@{}}
\toprule[1.2pt]
\multirow{2}{*}{Dataset} & Step     & \multicolumn{3}{c|}{5}   & \multicolumn{3}{c|}{10}  & \multicolumn{3}{c|}{15}  & \multicolumn{3}{c}{20}   \\ \cmidrule(l){2-2} \cmidrule(l){3-5} \cmidrule(l){6-8} \cmidrule(l){9-11} \cmidrule(l){12-14}
                         & Metric   & AUC$\uparrow$    & ACC$\uparrow$    & RMSE$\downarrow$   & AUC    & ACC    & RMSE   & AUC    & ACC    & RMSE   & AUC    & ACC    & RMSE   \\ \midrule
\multirow{12}{*}{EdNet}  & DKT      & 0.6767 & 0.6406 & 0.4705 & 0.6724 & 0.6363 & 0.4721 & 0.6688 & 0.6333 & 0.4731 & 0.6669      & 0.6322      & 0.4738      \\ \cmidrule(l){2-14} 
                         & SAKT     & 0.6737 & 0.6389 & 0.4718 & 0.6706 & 0.6370 & 0.4729 & 0.6661 & 0.6334 & 0.4741 & 0.6662      & 0.6328      & 0.4741      \\
                         & AKT      & 0.6793 & 0.6388 & 0.4703 & 0.6750 & 0.6366 & 0.4717 & 0.6719 & 0.6333 & 0.4728 & 0.6699      & 0.6316      & 0.4735      \\ \cmidrule(l){2-14} 
                         & Mamba4KT & 0.6655 & 0.6350 & 0.4753 & 0.6632 & 0.6312 & 0.4753 & 0.6609 & 0.6288 & 0.4763 & 0.6586      & 0.6273      & 0.4769      \\ \cmidrule(l){2-14} 
                         & DKVMN    & 0.6709 & 0.6366 & 0.4722 & 0.6682 & 0.6341 & 0.4731 & 0.6641 & 0.6319 & 0.4742 & 0.6626      & 0.6308      & 0.4747      \\ \cmidrule(l){2-14} 
                         & ATKT     & 0.6704 & 0.6371 & 0.4735 & 0.6669 & 0.6355 & 0.4747 & 0.6633 & 0.6323 & 0.4758 & 0.6625      & 0.6305      & 0.4762      \\ \cmidrule(l){2-14} 
                         & CL4KT    & -      & -      & -      & -      & -      & -      & -      & -      & -      & -      & -      & -      \\ \cmidrule(l){2-14} 
                         & Deep-IRT & 0.6573 & 0.6250 & 0.4792 & 0.6546 & 0.6216 & 0.4808 & 0.6499 & 0.6185 & 0.4820 & 0.6467      & 0.6170      & 0.4816      \\
                         & AT-DKT   & \underline{0.6816} & \underline{0.6442} & \underline{0.4693} & \underline{0.6787} & \underline{0.6414} & \underline{0.4703} & \underline{0.6752} & \underline{0.6375} & \underline{0.4716} & \underline{0.6722}      & \underline{0.6355}      & \textbf{0.4727}      \\ \cmidrule(l){2-14} 
                         & {\cellcolor[HTML]{CBCEFB}\textbf{DKT2}}     & {\cellcolor[HTML]{CBCEFB}\textbf{0.6853}} & {\cellcolor[HTML]{CBCEFB}\textbf{0.6451}} & {\cellcolor[HTML]{CBCEFB}\textbf{0.4683}} & {\cellcolor[HTML]{CBCEFB}\textbf{0.6809}} & {\cellcolor[HTML]{CBCEFB}\textbf{0.6420}} & {\cellcolor[HTML]{CBCEFB}\textbf{0.4697}} & {\cellcolor[HTML]{CBCEFB}\textbf{0.6771}} & {\cellcolor[HTML]{CBCEFB}\textbf{0.6379}} & {\cellcolor[HTML]{CBCEFB}\textbf{0.4709}} & {\cellcolor[HTML]{CBCEFB}\textbf{0.6731}}      & {\cellcolor[HTML]{CBCEFB}\textbf{0.6360}}      & {\cellcolor[HTML]{CBCEFB}\underline{0.4732}}      \\ \midrule
\multirow{12}{*}{Comp}   & DKT      & 0.7419 & 0.8097 & \underline{0.3722} & 0.7303 & 0.8086 & 0.3745 & 0.7208 & 0.8085 & 0.3759 & 0.7128      & 0.8089      & \underline{0.3765}      \\ \cmidrule(l){2-14} 
                         & SAKT     & 0.7418 & \underline{0.8098} & 0.3725 & 0.7307 & \underline{0.8087} & 0.3746 & 0.7213 & \underline{0.8086} & 0.3760 & 0.7130      & \textbf{0.8092}      & 0.3766      \\
                         & AKT      & 0.7384 & 0.8081 & 0.3737 & 0.7262 & 0.8069 & 0.3762 & 0.7213 & \underline{0.8086} & 0.3759 & 0.7073      & 0.8077      & 0.3782      \\ \cmidrule(l){2-14} 
                         & Mamba4KT & 0.7424 & 0.8097 & 0.3723 & 0.7310 & 0.8084 & 0.3746 & \textbf{0.7225} & 0.8085 & \underline{0.3757} & 0.7137      & 0.8090      & \underline{0.3765}      \\ \cmidrule(l){2-14} 
                         & DKVMN    & 0.7397 & 0.8091 & 0.3729 & 0.7286 & 0.8084 & 0.3750 & 0.7201 & 0.8082 & 0.3762 & 0.7117      & 0.8089      & 0.3768      \\ \cmidrule(l){2-14} 
                         & ATKT     & 0.7405 & 0.8094 & 0.3726 & 0.7302 & 0.8086 & 0.3746 & 0.7192 & 0.8082 & 0.3763 & 0.7103      & 0.8087      & 0.3771      \\ \cmidrule(l){2-14} 
                         & CL4KT    & 0.7364 & 0.8070 & 0.3746 & \textbf{0.7339} & 0.8082 & \textbf{0.3743} & 0.7142 & 0.8066 & 0.3778 & \textbf{0.7184}      & 0.8082      & \textbf{0.3763}      \\ \cmidrule(l){2-14} 
                         & Deep-IRT & 0.7372 & 0.8084 & 0.3737 & 0.7255 & 0.8075 & 0.3760 & 0.7159 & 0.8073 & 0.3773 & 0.7072      & 0.8079      & 0.3780      \\
                         & AT-DKT   & \underline{0.7440} & \underline{0.8098} & \textbf{0.3718} & 0.7311 & 0.8086 & \underline{0.3744} & 0.7212 & 0.8083 & 0.3758 & 0.7123      & \underline{0.8091}      & 0.3767      \\ \cmidrule(l){2-14}
                         & \textbf{{\cellcolor[HTML]{CBCEFB}DKT2}}     & {\cellcolor[HTML]{CBCEFB}\textbf{0.7459}} & {\cellcolor[HTML]{CBCEFB}\textbf{0.8103}} & {\cellcolor[HTML]{CBCEFB}\underline{0.3722}} & {\cellcolor[HTML]{CBCEFB}\underline{0.7328}} & {\cellcolor[HTML]{CBCEFB}\textbf{0.8089}} & {\cellcolor[HTML]{CBCEFB}0.3746} & {\cellcolor[HTML]{CBCEFB}\underline{0.7219}} & {\cellcolor[HTML]{CBCEFB}\textbf{0.8093}} & {\cellcolor[HTML]{CBCEFB}\textbf{0.3753}} & {\cellcolor[HTML]{CBCEFB}\underline{0.7152}}      & {\cellcolor[HTML]{CBCEFB}0.8090}      & {\cellcolor[HTML]{CBCEFB}\underline{0.3765}}      \\ \bottomrule[1.2pt]
\end{tabular}
}
\caption{Multi-step prediction performance of DKT2 and several representative baselines on EdNet and Comp.}
\label{apx: apx-multi-step}
\vspace{-0.3cm}
\end{table*}

\begin{figure*}[t]
    \includegraphics[width=\linewidth]{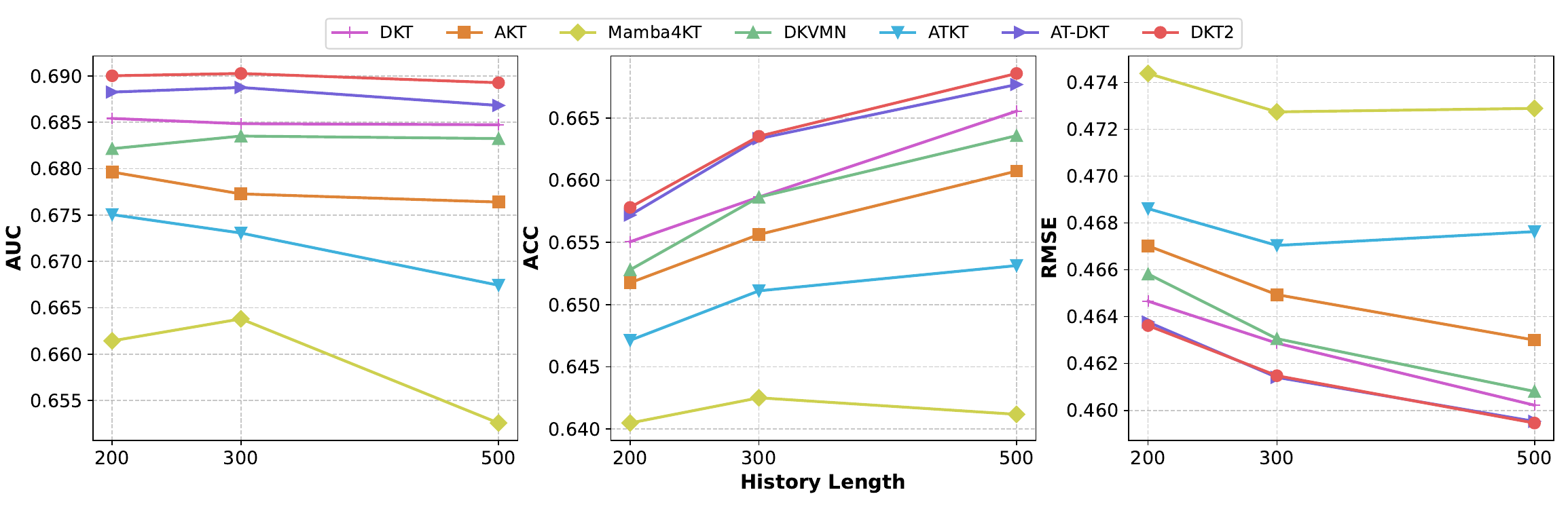}
    \vspace{-0.7cm}
    \caption{The prediction performance of DKT2 and several representative baselines on EdNet with different history lengths.}
    \label{fig: histroy_ednet}
    \vspace{-0.3cm}
\end{figure*}

\begin{figure*}[t]
    \includegraphics[width=\linewidth]{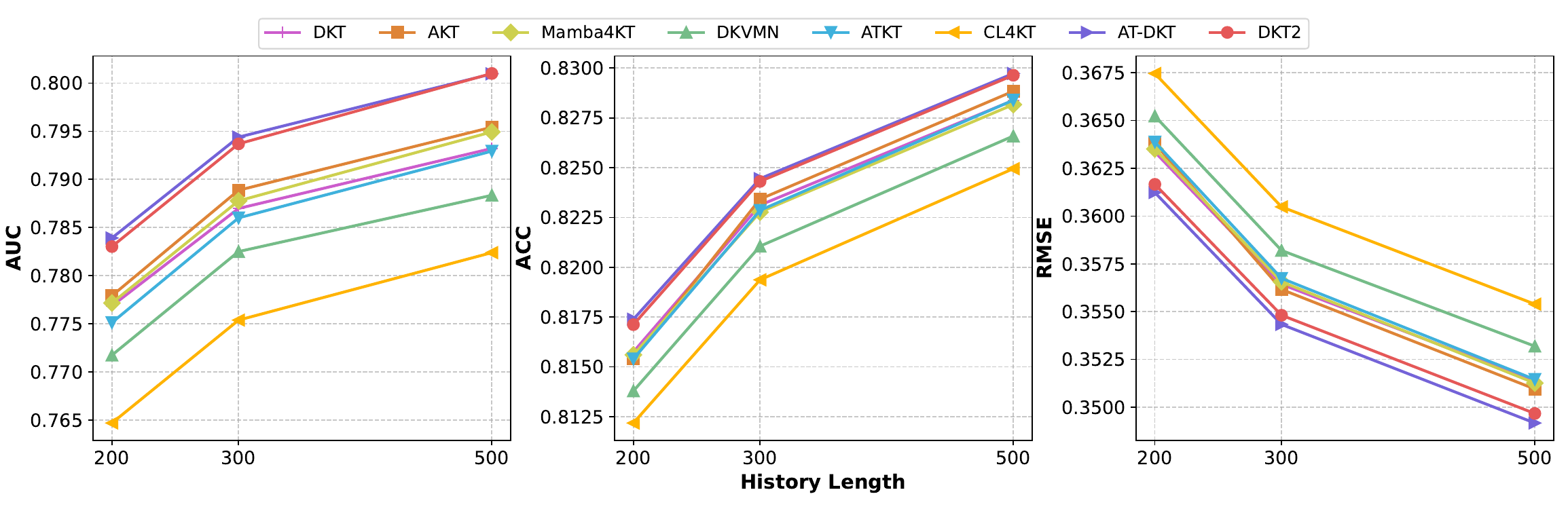}
    \vspace{-0.7cm}
    \caption{The prediction performance of DKT2 and several representative baselines on Comp with different history lengths.}
    \label{fig: histroy_comp}
    \vspace{-0.3cm}
\end{figure*}

\subsection{Baseline Description}
\label{apx: apx-baseline}
Here is a detailed description of the 18 baselines from 8 different categories in our experiment.

\begin{itemize}[leftmargin=*]
    \item{\textbf{Deep sequential models}}
    \begin{itemize}
        \item{\textbf{DKT}~\cite{NIPS2015_bac9162b}: DKT is a pioneering model that utilizes Recurrent Neural Networks (RNNs), specifically a single-layer Long Short-Term Memory (LSTM) network, to directly model students' learning processes and predict their performance.}
        \item{\textbf{DKT+}~\cite{yeung2018addressing}: DKT+ is an enhanced version of DKT. It addresses the reconstruction and prediction inconsistency issues present in the DKT by introducing additional regularization terms to the loss function.}
        \item{\textbf{DKT-F}~\cite{nagatani2019augmenting}: DKT-F improves upon DKT by incorporating students' forgetting behaviors into the modeling process.}
    \end{itemize}

    \item{\textbf{Attention-based models}}
    \begin{itemize}
        \item{\textbf{SAKT}~\cite{pandey2019self}: SAKT leverages self-attention networks to analyze and understand the complex relationships between concepts and a student's historical interactions with learning materials.}
        \item{\textbf{AKT}~\cite{ghosh2020context}: AKT is an advanced KT model that incorporates a Rasch model to regularize concept and question embeddings and a modified Transformer architecture with adaptive attention weights computed by a distance-aware exponential decay to account for the time distance between questions and students' previous interactions.}
        \item{\textbf{simpleKT}~\cite{liu2023simplekt}: simpleKT is a simple but tough-to-beat baseline to KT that combines simplicity with robust performance.}
        \item{\textbf{FoLiBiKT}~\cite{im2023forgetting}: FoLiBi enhances attention-based KT models by incorporating a forgetting-aware linear bias mechanism. We introduce FoLiBi with AKT, namely FoLiBiKT.}
        \item{\textbf{sparseKT}~\cite{huang2023towards}: sparseKT employs a k-selection module with soft-thresholding sparse attention (sparseKT-soft) and top-K sparse attention (sparseKT-topK) to focus on high-attention items, ensuring efficient and focused attention on the most relevant items.}
        \item{\textbf{DTransformer}~\cite{yin2023tracing}: DTransformer integrates question-level mastery with knowledge-level diagnosis through the use of Temporal and Cumulative Attention (TCA) and multi-head attention mechanisms. Additionally, a contrastive learning-based algorithm is used for enhancing the stability of the knowledge state diagnosis process.}
        \item{\textbf{stableKT}~\cite{li2024enhancing}: stableKT excels in length generalization, delivering stable and consistent performance across both short and long student interaction sequences. It employs a multi-head aggregation module that integrates dot-product and hyperbolic attention to capture hierarchical relationships between questions and their associated concepts.}
    \end{itemize}

    \item{\textbf{Mamba-based models}}
    \begin{itemize}
        \item{\textbf{Mamba4KT}~\cite{cao2024mamba4kt}: By leveraging Mamba, a state-space model supporting parallelized training and linear-time inference, Mamba4KT achieves efficient resource utilization, balancing time and space consumption.}
    \end{itemize}

    \item{\textbf{Graph-based models}}
    \begin{itemize}
        \item{\textbf{GKT}~\cite{nakagawa2019graph}: GKT revolutionizes the traditional KT task by employing Graph Neural Networks (GNNs) to represent the relationships between concepts as a graph.}
    \end{itemize}

    \item{\textbf{Memory-augmented models}}
    \begin{itemize}
        \item{\textbf{DKVMN}~\cite{zhang2017dynamic}: DKVMN employs a static key matrix to capture the interrelationships among latent concepts and a dynamic value matrix for continuously updating and predicting a student's knowledge mastery in real-time.}
        \item{\textbf{SKVMN}~\cite{abdelrahman2019knowledge}: SKVMN combines recurrent modeling of DKT with memory networks of DKVMN to enhance tracking of learners' knowledge states over time.}
    \end{itemize}
    
    \item{\textbf{Adversarial-based models}}
    \begin{itemize}
        \item{\textbf{ATKT}~\cite{guo2021enhancing}: ATKT is an attention-based LSTM model that employs adversarial training techniques to enhance generalization and reduce overfitting by applying perturbations to student interaction sequences.}
    \end{itemize}

    \item{\textbf{Contrastive learning-based models}}
    \begin{itemize}
        \item{\textbf{CL4KT}~\cite{lee2022contrastive}: CL4KT employs contrastive learning on augmented learning histories to enhance representation learning by distinguishing between similar and dissimilar student learning patterns.}
    \end{itemize}

    \item{\textbf{Other representative models}}
    \begin{itemize}
        \item{\textbf{Deep-IRT}~\cite{yeung2019deep}: Deep-IRT is an explainable KT model that combines the DKVMN with Item Response Theory (IRT) to provide detailed insights into learner trajectories and concept difficulties, bridging deep learning capabilities with psychometric interpretability.}
        \item{\textbf{AT-DKT}~\cite{liu2023enhancing}: AT-DKT enhances the original DKT by incorporating two auxiliary learning tasks: one focused on predicting question tags and the other on evaluating individualized prior knowledge.}
    \end{itemize}
\end{itemize}

\begin{table*}[]
\centering
\resizebox{\textwidth}{!}{%
\begin{tabular}{@{}c|c|c|c|c|c|c|c|c|c|c|c@{}}
\toprule[1.2pt]
\multicolumn{1}{c|}{Datasets} & \multicolumn{1}{c|}{Settings}                & Metrics & AKT    & simpleKT & FoLiBiKT & sparseKT & DTransformer & stableKT & DKVMN  & CL4KT  & Deep-IRT \\ \midrule 
\multirow{9}{*}{EdNet}        &   \multicolumn{1}{|c|}{\cellcolor{gray!10}}                                & AUC$\uparrow$     & 0.6083 & 0.6218   & 0.6098   & 0.6210   & 0.6140    & 0.6212   & 0.6195 & -      & 0.6190   \\
                              & \multicolumn{1}{|c|}{\cellcolor{gray!10}$\text{\tiny $\vartriangle$}$} & ACC$\uparrow$     & 0.5883 & 0.5938   & 0.5886   & 0.5916   & 0.5882    & 0.5907    & 0.5916 & -      & 0.5896   \\
                              &  \multicolumn{1}{|c|}{\cellcolor{gray!10}}                                  & RMSE$\downarrow$    & 0.4900 & 0.4896   & 0.4906   & 0.4901   & 0.4960   & 0.4894     & 0.4886 & -      & 0.4892   \\ \cmidrule(l){2-11} 
                              &  \multicolumn{1}{|c|}{\cellcolor{gray!10}}                              & AUC$\uparrow$     & 0.6806 & 0.6903   & 0.6794   & 0.6853   & 0.6717   & 0.6775     & 0.6564 & -      & 0.6577   \\
                              & \multicolumn{1}{|c|}{\cellcolor{gray!10}$\circ$}                                      & ACC$\uparrow$     & 0.6309 & 0.6388   & 0.6325   & 0.6341   & 0.6258    & 0.6286    & 0.6138 & -      & 0.6149   \\
                              &  \multicolumn{1}{|c|}{\cellcolor{gray!10}}                                & RMSE$\downarrow$    & 0.4798 & 0.4774   & 0.4815   & 0.4804   & 0.4895    & 0.4838    & 0.4815 & -      & 0.4815   \\ \cmidrule(l){2-11} 
                              &  \multicolumn{1}{|c|}{\cellcolor{gray!10}}                          & AUC$\uparrow$     & 0.6770 & 0.6832   & 0.6761   & 0.6813    & 0.6705    & 0.6801    & 0.6559 & -      & 0.6559   \\
                              & \multicolumn{1}{|c|}{\cellcolor{gray!10}$\bullet$}                                    & ACC$\uparrow$     & 0.6303 & 0.6324   & 0.6291   & 0.6336   & 0.6245    & 0.6291    & 0.6141 & -      & 0.6140   \\
                              & \multicolumn{1}{|c|}{\cellcolor{gray!10}}                         & RMSE$\downarrow$    & 0.4824 & 0.4804   & 0.4879   & 0.4795   & 0.4925    & 0.4849    & 0.4818 & -      & 0.4820   \\ \midrule
\multirow{9}{*}{Comp}         & \multicolumn{1}{|c|}{\cellcolor{gray!10}}                             & AUC$\uparrow$     & 0.7223 & 0.7208   & 0.7224   & 0.7171   & 0.7198     & 0.7194   & 0.7170 & 0.7184 & 0.7157   \\
                              & \multicolumn{1}{|c|}{\cellcolor{gray!10}$\text{\tiny $\vartriangle$}$} & ACC$\uparrow$     & 0.7887 & 0.7873   & 0.7877   & 0.7857   & 0.7806    & 0.7866    & 0.7850 & 0.7852 & 0.7842   \\
                              & \multicolumn{1}{|c|}{\cellcolor{gray!10}}                            & RMSE$\downarrow$    & 0.3919 & 0.3923   & 0.3921   & 0.3933   & 0.3955    & 0.3934    & 0.3938 & 0.3938 & 0.3946   \\ \cmidrule(l){2-11} 
                              &  \multicolumn{1}{|c|}{\cellcolor{gray!10}}                           & AUC$\uparrow$     & 0.8252 & 0.8251   & 0.8246   & 0.8217   & 0.8192     & 0.8160   & 0.7544 & 0.7465 & 0.7523   \\
                              & \multicolumn{1}{|c|}{\cellcolor{gray!10}$\circ$}                                      & ACC$\uparrow$     & 0.8146 & 0.8145   & 0.8157   & 0.8165   & 0.8165    & 0.8115    & 0.7932 & 0.7920 & 0.7922   \\
                              &  \multicolumn{1}{|c|}{\cellcolor{gray!10}}                                & RMSE$\downarrow$    & 0.3612 & 0.3612   & 0.3607   & 0.3612   & 0.3621    & 0.3644    & 0.3837 & 0.3863 & 0.3845   \\ \cmidrule(l){2-11} 
                              &  \multicolumn{1}{|c|}{\cellcolor{gray!10}}                              & AUC$\uparrow$     & 0.8160 & 0.8206   & 0.8157   & 0.8153   & 0.8107   & 0.8243     & 0.7450 & 0.7592 & 0.7430   \\
                              & \multicolumn{1}{|c|}{\cellcolor{gray!10}$\bullet$}                                    & ACC$\uparrow$     & 0.8097 & 0.8149   & 0.8160   & 0.8146   & 0.8118    & 0.8139    & 0.7907 & 0.7933 & 0.7897   \\
                              & \multicolumn{1}{|c|}{\cellcolor{gray!10}}                        & RMSE$\downarrow$    & 0.3664 & 0.3617   & 0.3621   & 0.3632   & 0.3662    & 0.3616    & 0.3866 & 0.3830 & 0.3871   \\ \bottomrule[1.2pt]
\end{tabular}
}
\caption{The prediction performance of KT models with weak applicability and comprehensiveness in the last 5 steps on EdNet and Comp under three different input settings. The $\text{\tiny $\vartriangle$}$ setting represents masking all interaction information (including questions, concepts and responses) for the last 5 steps, the $\circ$ setting represents masking the responses for the last 5 steps, without masking questions and concepts, and the $\bullet$ setting represents no masking, \textit{i.e.}, predicting the responses under the regular setting.}
\label{apx: apx-settings}
\vspace{-0.3cm}
\end{table*}

\subsection{Additional Experimental Results}
\label{apx: experiment}

\begin{figure*}[t]
\includegraphics[width=\linewidth]{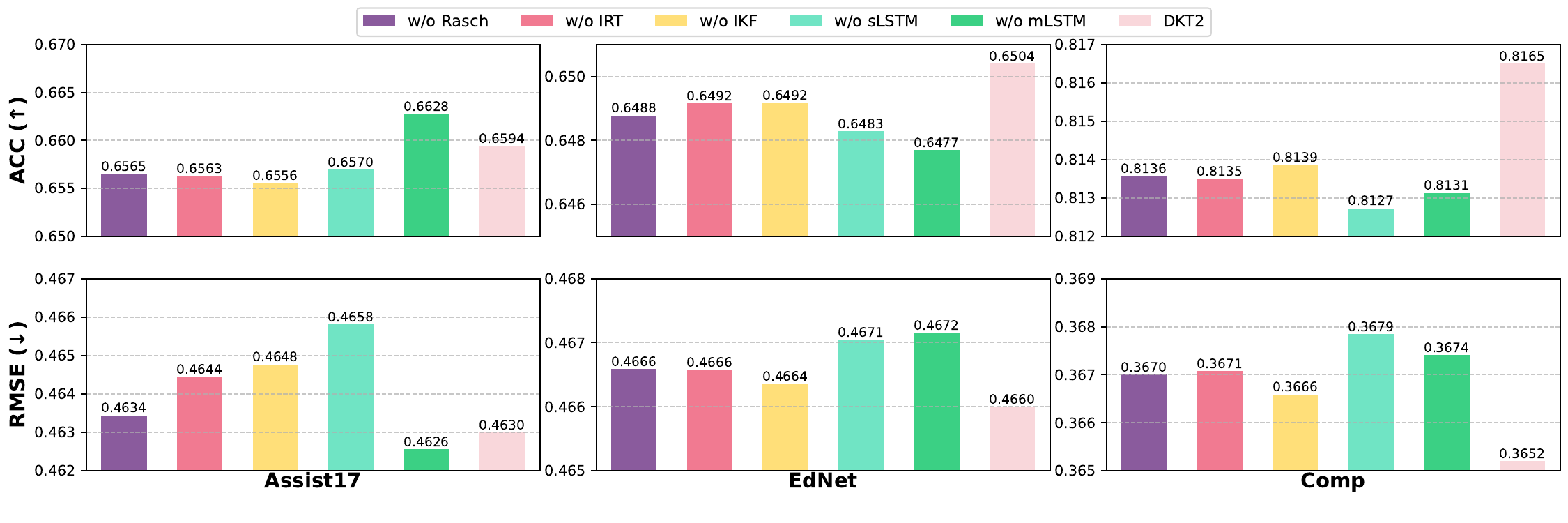}
    \vspace{-0.9cm}
    \caption{Ablation study on ACC and RMSE.}
    \label{fig: ablation_acc_rmse}
    \vspace{-0.5cm}
\end{figure*}

Due to space limitations in the main text, we have supplemented some additional experimental results here, including:
\begin{itemize}[leftmargin=*]
    \item{Multi-step prediction results on EdNet and Comp datasets;}
    \item{Prediction results with varying history lengths on EdNet and Comp;}
    \item{Prediction results under three different input settings on EdNet and Comp;}
    \item{Ablation study on ACC and RMSE.}
\end{itemize}

\subsubsection{Multi-step Prediction Results}
\label{apx: multi-step-experiment}
Table~\ref{apx: apx-multi-step} shows the multi-step (step=5, 10, 15, 20) prediction performance of DKT2 and several representative baselines from different categories on EdNet and Comp.

\subsubsection{Prediction Results with Varying History Lengths}
\label{apx: varying-length-experiment}
Fig.~\ref{fig: histroy_ednet} and Fig.~\ref{fig: histroy_comp} show the prediction performance of DKT2 and several representative baselines with different history lengths on the EdNet and Comp, respectively.

\subsubsection{Prediction Results Under Three Input Settings}
\label{apx: input-setting-experiment}
Table~\ref{apx: apx-settings} presents the prediction performance of KT models with weak applicability and comprehensiveness in the last 5 steps on EdNet and Comp under three different input settings.

\subsubsection{Ablation Study on ACC and RMSE}
\label{apx: ablation}

Fig.~\ref{fig: ablation_acc_rmse} shows the ablation study of DKT2 on ACC and RMSE.

%
%
%
%

\end{document}